\documentclass{article}

\usepackage{arxiv}
\usepackage{bm}
\usepackage[utf8]{inputenc} % allow utf-8 input
\usepackage[T1]{fontenc}    % use 8-bit T1 fonts
\usepackage{hyperref}       % hyperlinks
\usepackage{url}            % simple URL typesetting
\usepackage{booktabs}       % professional-quality tables
\usepackage{amsfonts}       % blackboard math symbols
\usepackage{nicefrac}       % compact symbols for 1/2, etc.
\usepackage{microtype}      % microtypography
\usepackage{lipsum}
\usepackage{adjustbox}
\usepackage{amsmath,amsthm,amssymb,amsfonts} % this for /equation* (without numbering)
\usepackage{algorithm}
\usepackage[noend]{algpseudocode}
\usepackage{soul}
\usepackage{booktabs,caption}
\usepackage[flushleft]{threeparttable}
\usepackage[title]{appendix}

\usepackage{array}
\usepackage{makecell}

\usepackage[utf8]{inputenc}
 
\usepackage{listings}
\usepackage{color}
\definecolor{codegreen}{rgb}{0,0.6,0}
\definecolor{codegray}{rgb}{0.5,0.5,0.5}
\definecolor{codepurple}{rgb}{0.58,0,0.82}
\definecolor{backcolour}{rgb}{0.95,0.95,0.92}
\lstdefinestyle{mystyle}{
    backgroundcolor=\color{backcolour},   
    commentstyle=\color{codegreen},
    keywordstyle=\color{magenta},
    numberstyle=\tiny\color{codegray},
    stringstyle=\color{codepurple},
    basicstyle=\footnotesize,
    breakatwhitespace=false,         
    breaklines=true,                 
    captionpos=b,                    
    keepspaces=true,                 
    numbers=left,                    
    numbersep=5pt,                  
    showspaces=false,                
    showstringspaces=false,
    showtabs=false,                  
    tabsize=2,
    escapeinside={<@}{@>},
}
 
\usepackage{booktabs} % For \toprule, \midrule and \bottomrule
\usepackage{siunitx} % Formats the units and values

\usepackage{mathtools} % for arrows
\usepackage{gensymb} % To write degree C 
\usepackage{mhchem} % To write chemical symbols
\usepackage{fixltx2e} % To write subscripts and superscripts 
\usepackage{bbm}

\usepackage{multirow}

\usepackage{fancyhdr} % customize the footer and header
\theoremstyle{definition}

\theoremstyle{definition}

\theoremstyle{remark}

\usepackage{soul} %To highlight the text 
\soulregister\cite7
\soulregister\ref7
\soulregister\pageref7
\usepackage{bm}

\usepackage{framed} % Framing content

\usepackage{multicol} % Multiple columns environment

\usepackage{nomencl} % Nomenclature package
\usepackage{tikz}
\makenomenclature
\usepackage[resetlabels,labeled]{multibib}

\renewcommand*\nompreamble{\begin{multicols}{2}}

\renewcommand*\nompostamble{\end{multicols}}

\usepackage{amssymb}% http://ctan.org/pkg/amssymb
\usepackage{pifont}% http://ctan.org/pkg/pifont
\newcommand{\cmark}{\ding{52}}%
\newcommand{\xmark}{\ding{53}}%
\definecolor{light-gray}{gray}{0.95}
\newcommand{\code}[1]{\colorbox{light-gray}{\small\color{blue}\textbf{\texttt{#1}}}}

\soulregister\cite7
\soulregister\ref7
\soulregister\pageref7
\usepackage{bm}
\usepackage{pdfpages}

\usepackage[]{hyperref}
\hypersetup{
  colorlinks   = true, %Colours links instead of ugly boxes
  urlcolor     = blue, %Colour for external hyperlinks
  linkcolor    = blue, %Colour of internal links
  citecolor   = blue %Colour of citations
}

\title{NEORL: \textbf{N}euro\textbf{E}volution \textbf{O}ptimization with \textbf{R}einforcement \textbf{L}earning}

\author{
Majdi I. Radaideh\thanks{Review in Progress $^{\bigstar}$Authors with equal contribution} \\
Department of Nuclear Science and Engineering,\\
Massachusetts Institute of Technology,\\
Cambridge, MA 02139, United States \\
\texttt{radaideh@mit.edu} \\
\And
Katelin Du$^{\bigstar}$\\
Department of Nuclear Science and Engineering,\\
Massachusetts Institute of Technology,\\
Cambridge, MA 02139, United States \\
\And
Paul Seurin$^{\bigstar}$\\
Department of Nuclear Science and Engineering,\\
Massachusetts Institute of Technology,\\
Cambridge, MA 02139, United States \\
\And
Devin Seyler\\
Department of Physics, \\
Massachusetts Institute of Technology, \\
Cambridge, MA 02142, United States\\
\And
Xubo Gu\\
School of Nuclear Science and Engineering, \\
Shanghai Jiao Tong University, \\
Shanghai 200240, China
\And
Haijia Wang\\
Department of Electrical Engineering and Computer Science, \\
Massachusetts Institute of Technology, \\
Cambridge, MA 02142, United States
\And
Koroush Shirvan \\
Department of Nuclear Science and Engineering,\\
Massachusetts Institute of Technology,\\
Cambridge, MA 02139, United States \\
}

\begin{document}

\maketitle

\begin{abstract}
We present an open-source Python framework for NeuroEvolution Optimization with Reinforcement Learning (NEORL) developed at the Massachusetts Institute of Technology. NEORL offers a global optimization interface of state-of-the-art algorithms in the field of evolutionary computation, neural networks through reinforcement learning, and hybrid neuroevolution algorithms. NEORL features diverse set of algorithms, user-friendly interface, parallel computing support, automatic hyperparameter tuning, detailed documentation, and demonstration of applications in mathematical and real-world engineering optimization. NEORL encompasses various optimization problems from combinatorial, continuous, mixed discrete/continuous, to high-dimensional, expensive, and constrained engineering optimization. NEORL is tested in variety of engineering applications relevant to low carbon energy research in addressing solutions to climate change. The examples include nuclear reactor control and fuel cell power production. The results demonstrate NEORL competitiveness against other algorithms and optimization frameworks in the literature, and a potential tool to solve large-scale optimization problems. More examples and benchmarking of NEORL can be found here: \href{https://neorl.readthedocs.io/en/latest/index.html}{https://neorl.readthedocs.io/en/latest/index.html}
\end{abstract}

% keywords can be removed
\keywords{Optimization \and Deep Reinforcement Learning \and Evolutionary Computation \and Neuroevolution \and Nuclear Reactor Design, Carbon-free energy}

\section{Introduction}
\label{sec:intro}

Optimization refers to the process of finding the best possible solution(s) such that an objective function is met. As the real-world engineering designs grow in complexity, the needs for new optimization techniques and friendly frameworks to implement them become more needed than before. In recent decades, global optimization techniques have been extensively applied to various domains, including but not limited to engineering \cite{stewart2021survey,rajput2020review}, machine learning \cite{bennett2006interplay}, finance \cite{soler2017survey}, health care \cite{ahmadi2017outpatient}, and many others. In this literature review, we classify optimization algorithms into four major categories relevant to our proposed research: evolutionary algorithms (EA), neural-based algorithms, hybrid neuroevolution, and gradient-based algorithms. 

Genetic algorithms (GA) \cite{sivanandam2008genetic} are undoubtedly the face and origin of a long chain of metaheutrstics and EAs. GA, which evolve population using natural operators (e.g. crossover, mutation, selection), inspired a set of algorithms called ``evolution strategies'' (ES). ES are effective for continuous optimization \cite{glasmachers2010exponential, hansen2006cma} and have been used as an alternative to gradient descent to train neural networks \cite{salimans2017evolution}. Following GA footprints, a set of classical evolutionary and swarm algorithms has been developed including particle swarm optimization (PSO) \cite{kennedy1995particle}, differential evolution (DE) \cite{storn1997differential}, ant colony optimizer (ACO) \cite{dorigo1999ant}, and tabu search (TS) \cite{glover1989tabu}, which dominated the field until the late 1990s. During the early 2000s, new EA started to emerge such as artificial bee colony (ABC) \cite{karaboga2007powerful}, cuckoo search (CS) \cite{yang2009cuckoo}, and bat algorithm (BAT) \cite{yang2010new}. In the past 10 years, the development of EA has become more nature-inspired. The main modern EAs that show promise and popularity are: grey wolf optimizer (GWO) \cite{mirjalili2014grey}, whale optimization algorithm (WOA) \cite{mirjalili2016whale}, moth-flame optimization (MFO) \cite{mirjalili2015moth}, salp swarm algorithm (SSA) \cite{mirjalili2017salp}, and recently the Harris hawks optimizer (HHO) \cite{heidari2019harris}. An overview of classical EA and evolution strategies can be found here \cite{slowik2020evolutionary}, while modern EAs and their applications are reviewed here \cite{mirjalili2019nature, faris2018grey}. Implementation-wise in Python, DEAP \cite{DEAP_JMLR2012} provides a solid framework for classical EA such as GA, genetic programming, and evolution strategies, while EvoloPy \cite{faris2016evolopy} provides an implementation of modern EA such as GWO, HHO, WOA, and MFO along with some basic implementation of classical GA and PSO. 

Although neural networks are probably older than EA, they started to emerge back in the mainstream when more computing power and data became available \cite{schmidhuber2015deep}. Artificial neural networks trained by supervised learning demonstrated some success in solving global optimization problems \cite{joya2002hopfield, vinyals2015pointer}, even though their practical implementation always faces major challenges. The dimensionality of the search space and the ``dynamic'' nature of the optimization problem are restrictions for generating a representative labeled training data for the neural networks. Therefore, in this work, we focus on a certain class of neural network algorithms trained by reinforcement learning (RL) \cite{mnih2015human}. What makes RL special for optimization is that it does not require a pre-generated labeled data (like unsupervised or supervised learning). Alternatively, the agent collects features and labels as needed, guided by a reward signal (e.g. fitness value) to meet a certain reward value. This analogy seems ideal for optimization, as the agent can adapt to the search region to learn to generate individuals that maximize/minimize a certain objective function, which acts as the reward signal. This inspiration has led to the term ``learning to optimize'' \cite{li2016learning}, which has been followed by many attempts to use RL and neural networks to solve optimization problems. See for example \cite{bello2016neural} on using RL with recurrent neural network policy for combinatorial optimization, or \cite{khalil2017learning} for using graph embedding and RL to solve combinatorial optimization over graphs, or \cite{radaideh2021physics} for using deep Q learning and proximal policy optimization for physics-informed optimization of nuclear fuel. A comprehensive overview of using machine learning and neural networks to solve combinatorial optimization is conducted by \cite{bengio2021machine}. To our knowledge, currently, there is no framework providing generic implementation of neural networks and RL for optimization purposes. However, OpenAI baselines \cite{baselines} and stable-baselines \cite{stable-baselines} provided an implementation of the state-of-the-art RL algorithms for applications on game-playing AI and autonomous control. These frameworks provided a solid base for us as we have built on their implementation to leverage our RL and neural optimizers.  

The field of neuroevolution started to grow simultaneously as the neural and evolutionary fields do to obtain the best of both worlds. Neuroevolution augmenting topologies (NEAT) \cite{stanley2002evolving} and its modified version, hypercube-based NEAT (HyperNEAT) \cite{stanley2009hypercube} are probably the most common in this area. In these hybrid algorithms, GA is used to evolve neural network topolgies by using evolutionary operators to optimize network width, depth, weights, biases, and activation. NEAT has inspired other researchers to transfer neuroevolution to train deep neural networks and RL systems as shown by these GA and deep neuroevolution efforts \cite{such2017deep} or as shown by using PSO \cite{zhao2021particle, jiang2020efficient}. Neuroevolution can contribute to the field of optimization in two major forms. The first form is \textit{exchanging solutions between evolutionary and neural algorithms during search} to help increasing search diversity, improving solution quality, and accelerate convergence. Hybrid neuroevolution optimizers have been developed in the literature, for example, GA crossover and mutation have been controlled by Q-learning \cite{pettinger2002controlling}, and RL algorithms have been used to inform and guide EA search in constrained optimization \cite{radaideh2021rule}. Furthermore, EA are used as an alternative to temporal credit assignment and value function in RL \cite{moriarty1999evolutionary}. A comprehensive overview of RL, EA, and the history of their hybridization can be found in this review \cite{drugan2019reinforcement}. The second form is through \textit{using neural networks as surrogates to assist EA}. Although supervised neural networks have limited scope in tackling the optimization problem directly, supervised networks excelled as surrogate models to assist EA in tackling expensive fitness functions \cite{tenne2010computational}. A class of hybrid neuroevolution algorithms involves using EA to optimize the problem, while neural networks are trained based on evolutionary-generated data to approximate the original fitness function \cite{forrester2009recent}. For example, feedforward \cite{jin2002framework} and Bayesian \cite{briffoteaux2020parallel} neural network surrogates were used to assist EA in optimizing expensive functions. To our knowledge, currently, there is also no framework providing generic implementation of neuroevolution for optimization purposes either as an optimizer or as a hybrid surrogate algorithm.

Lastly, gradient-based or deterministic optimization algorithms exist in the literature and have been used for  decades. This optimization category offers fast search and convergence, but suffers from major issues such as local optima entrapment, requirement of derivation (i.e. closed-form fitness function is needed), and computational limitations to calculate the derivatives of high-dimensional problems. Indeed, these limitations encourage the development of the prescribed three optimization categories (neural, evolutionary, neuroevolution). Fortunately, gradient-based algorithms are well-implemented and tested. For example, nlopt \cite{johnson2014nlopt}, PyOpt \cite{perez2012pyopt}, and scipy.optimize \cite{virtanen2020scipy} are just few examples of robust Python implementation of popular algorithms such as BFGS, Newton-Conjugate-Gradient, Sequential Least Squares Programming (SLSQP), Method of Moving Asymptotes, Preconditioned truncated Newton, and a variety of trust-region-based methods. Given that gradient-based algorithms are mature enough with solid literature implementation, this category is not discussed further here. 

In this work, we propose NEORL (NeuroEvolution Optimization with Reinforcement Learning) as a new open-source Python framework that brings the latest developments of machine learning and evolutionary computation to serve the optimization research community, including energy and engineering applications. The novelty and uniqueness of NEORL compared to other optimization frameworks can be summarised as follows:
\begin{itemize}
    \item NEORL offer a diverse set of +25 algorithms that belong to evolutionary computation, neural networks, and neuroevolution categories. The diversity of the implemented algorithms allow the user to test different types of algorithms on the same problem; obtaining a more comprehensive understanding and better results. 
    \item NEORL includes efficient implementation of previously proposed algorithms. It also includes new algorithms proposed and developed by the authors themselves such as RL-informed evolutionary algorithms, EA with prioritized experience replay, and neural Harris hawks optimizer.  
    \item NEORL supports discrete, categorical, and continuous search spaces as well as mixed discrete-categorical-continuous optimization. We demonstrate how to use NEORL to solve the combinatorial travelling salesman problem, mixed-integer optimization, and pure continuous engineering optimization.
    \item NEORL supports parallel computing. Vast majority of NEORL algorithms offer parallel optimization to accelerate the optimization of expensive fitness functions, which can be simply activated by a flag. This highlights a major advantage of NEORL over other frameworks for large-scale expensive problems. 
    \item NEORL supports internal hyperparameter tuning. NEORL offers automatic search to tune the optimizer hyperparameters to maximize its performance. Some algorithms, especially neural-based, have multiple hyperparameters to tune. The user can choose between grid search, random search, evolutionary search, or Bayesian search for that purpose.
    \item NEORL is fully documented with a variety of unit tests, benchmarks, and engineering examples, which enable a friendly user experience. 
\end{itemize}

For the remaining sections of this work, the optimization methodology of major NEORL algorithms are described in section \ref{sec:method}. In section \ref{sec:struct}, NEORL structure is presented, which includes a step-by-step procedure to construct a complete NEORL problem, and a comparison with existing frameworks. Section \ref{sec:apps} presents NEORL application to variety of energy optimization problems that belong to nuclear power and fuel cell energy production. The reader can also find additional applications of NEORL in the webpage\footnote{\url{https://neorl.readthedocs.io/en/latest/index.html}}. Discussions and conclusions of this work are highlighted in section \ref{sec:conc}. 

\section{Optimization Methodology}
\label{sec:method}

Without loss of generality, the standard form of a minimization optimization problem can be formulated as:

\begin{equation}
\label{eq:single_opt}
\begin{aligned}
    &\underset {\vec{x}}{\text{min } } f(\vec{x}), \\
    &\text{subject to}, \\
    &g_i(\vec{x}) \geq 0, \quad i= 1, 2, ..., m, \\
    &h_j(\vec{x}) = 0, \quad j= 1, 2, ..., p, 
\end{aligned}
\end{equation}
where $f$ is the objective function to be minimized over the input vector $\vec{x}$, which has a size $d$. $g_i(\vec{x}) \leq 0$ are inequality constraints with a total of $m$ constraints, $h_j(\vec{x}) = 0$ are equality constraints with a total of $p$ constraints. When $p=0$ and $m=0$, the problem becomes unconstrained. The maximization problem can be formulated similarly by optimizing the negative form of the objective function $f$.

The discrete/combinatorial optimization problem formulation is similar to Eq.\eqref{eq:single_opt}, except that the values of $\vec{x}$ are restricted to discrete values (e.g. 1, 2, 3, ...). For the mixed discrete-continuous problems, the variable types in $\vec{x}$ can be different, e.g. $x_1$ is a continuous variable, $x_2$ is discrete, and so on. 

A summary of all supported optimization algorithms in NEORL is provided in Table \ref{tab:algs}, which includes their multiprocessing and input space support. In the next subsections, we discuss three different algorithm categories, which NEORL algorithms belong to. 

% Table generated by Excel2LaTeX from sheet 'algs'
\begin{table}[!h]
  \centering
  \small
  \caption{List of NEORL algorithms and their optimization support}
  \begin{threeparttable}
    \begin{tabular}{lllllll}
    \toprule
    Num.  & Algorithm & Category & \thead{Discrete \\ Space} & \thead{Continuous \\ Space} & \thead{Mixed \\ Space} & Multiprocessing \\
    \midrule
    1     & PPO   & Neural & \cmark & \cmark & \cmark & \cmark \\
    2     & A2C   & Neural & \cmark & \cmark & \cmark & \cmark \\
    3     & DQN   & Neural & \cmark & \xmark & \xmark & \xmark \\
    4     & ACKTR & Neural & \cmark & \cmark & \cmark & \cmark \\
    5     & ACER  & Neural & \cmark & \xmark & \xmark & \cmark \\
    6     & ES/GA & Evolutionary & \cmark & \cmark & \cmark & \cmark \\
    7     & PSO   & Evolutionary & \cmark & \cmark & \cmark & \cmark \\
    8     & DE    & Evolutionary & \cmark & \cmark & \cmark & \cmark \\
    9     & XNES  & Evolutionary & \xmark & \cmark & \xmark & \cmark \\
    10    & GWO   & Evolutionary & \cmark & \cmark & \cmark & \cmark \\
    11    & PESA  & Hybrid & \cmark & \cmark & \cmark & \cmark \\
    12    & PESA2 & Hybrid & \cmark & \cmark & \cmark & \cmark \\
    13    & SA    & Evolutionary* & \cmark & \cmark & \cmark & \cmark \\
    14    & SSA   & Evolutionary & \cmark & \cmark & \cmark & \cmark \\
    15    & WOA   & Evolutionary & \cmark & \cmark & \cmark & \cmark \\
    16    & HHO   & Evolutionary & \cmark & \cmark & \cmark & \cmark \\
    17    & MFO   & Evolutionary & \cmark & \cmark & \cmark & \cmark \\
    18    & JAYA  & Evolutionary & \cmark & \cmark & \cmark & \cmark \\
    19    & BAT   & Evolutionary & \cmark & \cmark & \cmark & \cmark \\
    20    & RNEAT & Hybrid & \xmark & \cmark & \xmark & \cmark \\
    21    & FNEAT & Hybrid & \xmark & \cmark & \xmark & \cmark \\
    22    & ACO   & Evolutionary & \xmark & \cmark & \xmark & \cmark \\
    23    & PPO-ES & Hybrid & \cmark & \cmark & \cmark & \cmark \\
    24    & ACKTR-DE & Hybrid & \cmark & \cmark & \cmark & \cmark \\
    25    & NHHO  & Hybrid & \cmark & \cmark & \cmark & \cmark \\
    26    & NGA  & Hybrid & \xmark & \cmark & \xmark & \xmark \\
    27    & CS  & Evolutionary & \cmark & \cmark & \cmark & \cmark \\
    28    & TS  & Evolutionary & \cmark & \xmark & \xmark & \xmark \\
    \bottomrule
    \end{tabular}%
    \begin{tablenotes}
      \small
      \item $*$ Simulated Annealing (SA) does not belong directly to evolutionary algorithms, but to a more generic category of stochastic optimization
    \end{tablenotes}
    \end{threeparttable}
  \label{tab:algs}%
\end{table}%

\subsection{Evolutionary Algorithms}
\label{sec:ea}

Also known as metaheuristics, EA in NEORL form a basis for direct optimization and also as a baseline to leverage hybrid neuroevolution algorithms. Almost all EA in NEORL follow the paradigm in Algorithm \ref{alg:ea}. A random population is generated, evaluated by a fitness function, best individuals are selected for further evolution, and the population is updated using natural operations (crossover, mutation, hunting strategies, food search, etc.) for the next generation. The main differences between EA are the types of natural operations and how the population individuals are communicating. For example, GA uses mutation and crossover, PSO uses cognitive and social speed to communicate, while GWO uses prey search, encircle, and hunting methods.  

\begin{algorithm}[!h]
  \small
    \caption{Pseudocode for evolutionary algorithms (Minimization)}
    \label{alg:ea}
    \begin{algorithmic}[1]
     \State \textbullet Set algorithm hyperparameters
     \State \textbullet Define the fitness function and input space
     \State \textbullet Initialize the population randomly with size $pop\_size$, i.e. $\{\vec{x}_1^0, \vec{x}_2^0, ..., \vec{x}_{pop\_size}^0\}$
     \State \textbullet Initialize $y_{best}$=$\infty$
     \For{Generation $k = 1$ to $N_{gen}$} 
             \For{Individual $j = 1$ to $pop\_size$}
                \State \textbullet Encoding-decoding tools are applied to obtain the individual in real space ($\vec{x}_j^0 \xrightarrow{} \vec{x}_j$)
                \State \textbullet Individual $\vec{x}_j$ is evaluated by the fitness function, and fitness $y_j$ is returned
                 \If {$y_j < y_{best}$}
                    \State \textbullet  Set $y_{best} = y_j$
                    \State \textbullet  Set $\vec{x}_{best} = \vec{x}_j$
                 \EndIf
            \EndFor 
        \State \textbullet (Optional step) Select the top $m$ individuals to survive to the next generation, i.e. $m < pop\_size$ 
        \State \textbullet Update the population position using evolutionary operations to minimize fitness in the next generation
    \EndFor 
    \State \textbullet  Return $\vec{x}_{best}$ and $y_{best}$
\end{algorithmic}
\end{algorithm}

We will highlight the Harris Hawks optimization (HHO) \cite{heidari2019harris} as one of the most recent EA. HHO employs different approaches to update the Hawks positions, which are chasing a prey (global optima) to hunt. After evaluating the first random generation, $\vec{x}^{rabbit}$ can be determined, which is the position of the closest hawk to the prey. In every subsequent generation, the following processes are applied for every hawk in the population. The energy of the prey is updated every generation using:
\begin{equation}
    E=2E_0(1-k/N_{gen}),
\end{equation}
where $E_0 = 2rand(0,1)-1$ is the baseline energy, $k$ is the current generation, and $N_{gen}$ is the total number of generations. Now if the absolute value of rabbit energy is more than 1, $|E| \geq 1$, the hawks activate the exploration phase expressed by:  
\begin{equation}
 x_{k+1} =\left\{
\begin{array}{ll}
\label{eq:hho_explore}
      \vec{x}^{rand}_k - r_1 |\vec{x}^{rand}_k - 2r_2\vec{x}_k|  & q \geq 0.5 \\
       (\vec{x}^{rabbit}_k - \vec{x}_k^m) - r_3 (\vec{x}_{min} + r_4 (\vec{x}_{max}-\vec{x}_{min}))& q < 0.5\\
\end{array} 
\right. 
\end{equation}
where $\vec{x}_{k+1}$ is the updated position of the hawk in the next generation, $\vec{x}^{rabbit}_k$ is the position of rabbit, $\vec{x}_k$ is the current position of the hawk, and $r_1, r_2, r_3, r_4$, $q$ are uniform random numbers between [0,1], which are regenerated in each new generation. $\vec{x}_{min}$ and $\vec{x}_{max}$ are the upper and lower bounds of the variables, $\vec{x}^{rand}_k$ is a randomly selected hawk from the current population, and $\vec{x}^m_k$ is the average position of the current population of hawks. As the reader can notice, significant amounts of randomness can be seen in Eq.\eqref{eq:hho_explore}, which are supposed to improve the exploration of the hawks in finding a prey. Now if $|E| < 1$, the hawks activate their exploitation phase, which involves one of four strategies. If $rand(0,1) \geq 0.5$ and $|E| \geq 0.5$, hawks perform soft besiege of the prey by
\begin{equation}
  \vec{x}_{k+1} = (\vec{x}^{rabbit}_k-\vec{x}_k) - E |J\vec{x}^{rabbit}_k-\vec{x}_k|,  
\end{equation}
where $J = 2(1 - rand(0,1))$ represents the random jump strength of the rabbit, which is assumed to be a random jump. Otherwise, if $rand(0,1) \geq 0.5$ and $|E| < 0.5$, hard besiege is applied as follows:
\begin{equation}
  \vec{x}_{k+1} = \vec{x}^{rabbit}_k - E |\vec{x}^{rabbit}_k-\vec{x}_k|.
\end{equation}

For brevity, the authors have added two more exploitation strategies where soft and hard besieges are conducted with progressive rapid dives toward the prey. Soft besiege with rapid dive occurs when \{$rand(0,1) < 0.5$ and $|E| \geq 0.5$\}, while hard besiege with rapid dive occurs when \{$rand(0,1) < 0.5$ and $|E| < 0.5$\}. The reader can refer to the original paper for more details about these exploitation strategies \cite{heidari2019harris}. Overall, after sufficient number of generations, continuous update of the hawk positions will lead to a solution close to the global optima or the prey. 

All EA in NEORL are completely or partially parallelized, where the population fitness is evaluated in parallel before the next position update. In particular, lines 6-11 of Algorithm \ref{alg:ea} can be executed in parallel, as these steps, especially line 8, are usually the most expensive parts of EA. All EA supported in NEORL are listed in Table \ref{tab:algs}, including their supported input space.  

\subsection{Neural Networks Algorithms}

\begin{figure}[h] 
 \centering
  \includegraphics[width=0.45\textwidth]{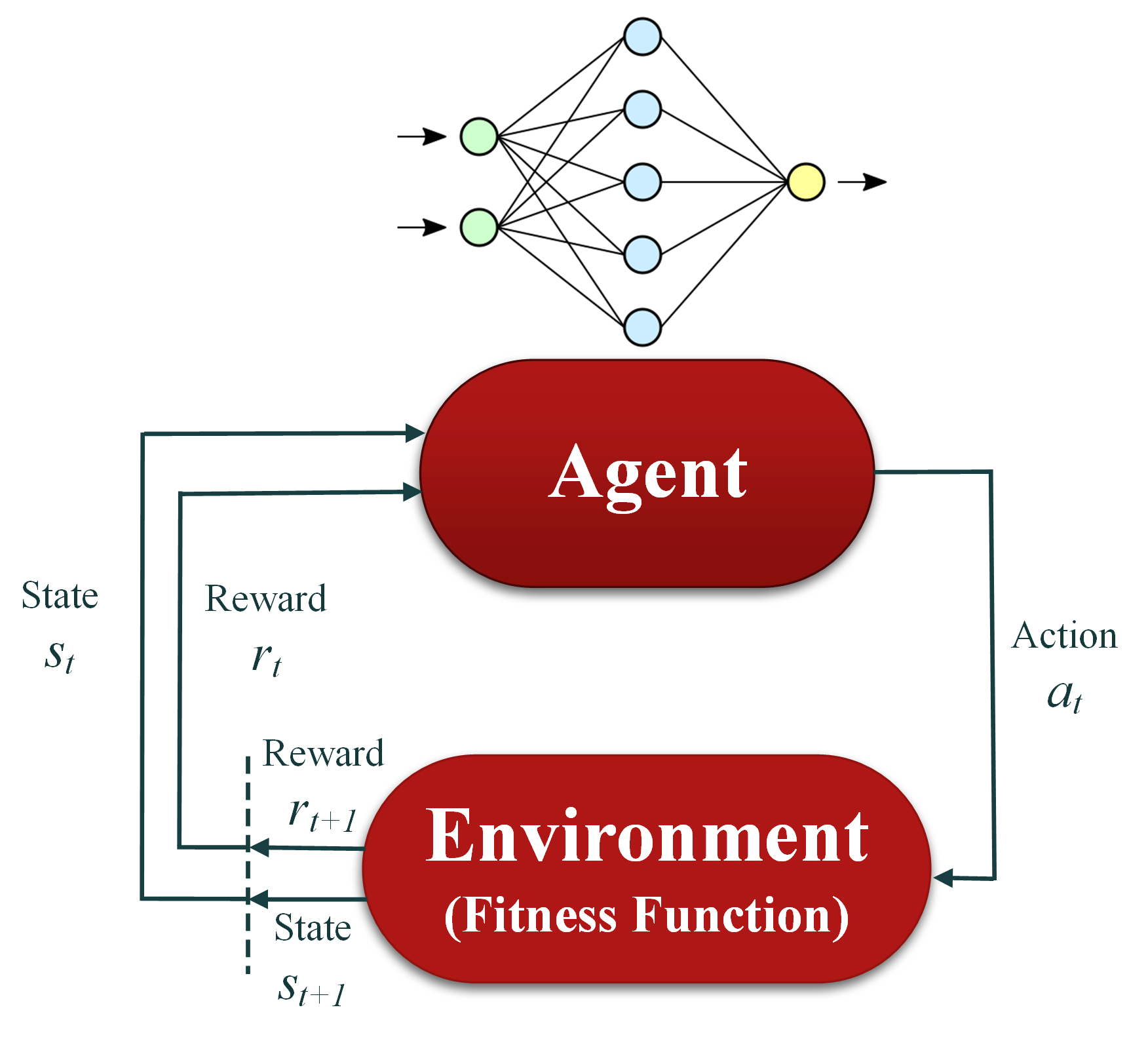}
  \caption{Reinforcement learning flow for optimization}
  \label{fig:rl}
\end{figure}

Reinforcement learning (RL) algorithms are the main neural-based algorithms in NEORL due to their natural fit to optimization as described before in section \ref{sec:intro}. Indeed, all RL algorithms in NEORL belong to the subcategory of deep RL, as deep neural networks are trained based on RL reward signal. However, for simplicity, we will use the RL acronym in this work. The following terms in EA (individual, population size, fitness, generation, evolutionary operator) are analogous to the following terms in RL (action, episode length, reward, episode, agent update). The RL paradigm flow is shown in Figure \ref{fig:rl} and described in Algorithm \ref{alg:rl}. The flow of RL and EA is quite similar with a major difference that the natural operators are replaced by a neural network model, which is trained continuously based upon the data collected from the fitness function. For optimization purposes, the action space ($a_t$) and state space ($s_t$) in Figure \ref{fig:rl} are identical (see also line 10 in Algorithm \ref{alg:rl}). Notice that for consistency with EA, in Algorithm \ref{alg:rl}, we refer to the action $a$ with symbol $\vec{x}$ to represent the EA input individual. RL algorithms are usually classified into value-based, policy gradient, and actor-critic, with the later category being the state-of-the-art.  

\begin{algorithm}[!h]
  \small
    \caption{Pseudocode for reinforcement learning algorithms (Maximization)}
    \label{alg:rl}
    \begin{algorithmic}[1]
     \State \textbullet Set algorithm hyperparameters
     \State \textbullet Initialize the environment (i.e. fitness function and input space)
     \State \textbullet Initialize the neural network parameters ($\theta$)
     \State \textbullet Initialize $y_{best}$=$-\infty$
     \For{Generation $k = 1$ to $N_{gen}$} 
             \For{Agent $j = 1$ to $pop\_size$}
                \State \textbullet Neural network proposes a base action ($\vec{x}_j^0$)
                \State \textbullet Encoding-decoding tools are applied to obtain the action in real space ($\vec{x}_j^0 \xrightarrow{} \vec{x}_j$)
                \State \textbullet Action $\vec{x}_j$ is evaluated by the environment, and fitness $y_j$ is returned as reward
                \State \textbullet State is set to current action, $\vec{s}_j \xleftarrow{}{} \vec{x}_j$ 
                 \If {$y_j > y_{best}$}
                    \State \textbullet  Set $y_{best} = y_j$
                    \State \textbullet  Set $\vec{x}_{best} = \vec{x}_j$
                 \EndIf
            \EndFor 
        \State \textbullet Update neural network parameters ($\theta$) to maximize fitness in the next generation
    \EndFor 
    \State \textbullet  Return $\vec{x}_{best}$ and $y_{best}$
\end{algorithmic}
\end{algorithm}

Q-learning or Quality-learning is the core of most value-based RL algorithms. In simple Q-learning, the Q value is updated recursively, as derived from the Bellman equation  

\begin{equation}
    Q^{new} (s_t, a_t) \xleftarrow{} (1-\alpha) \overbrace{Q (s_t, a_t)}^{\text{old value}} +  \underbrace{\alpha}_{\text{learning rate}} \overbrace{[\underbrace{r_t}_{\text{reward}} + \underbrace{\gamma}_{\text{discount factor}} \cdot \underbrace{\max \limits_a Q(s_{t+1}, a)}_{\text{optimum future value}}]}^{\text{learned value}},
\end{equation}
where $s_t$, $a_t$, and $s_{t+1}$ are the current state, current action, and next state, respectively. Q-learning requires saving all state-action pairs in a Q-table, which can be extremely large and expensive for complex problems. Alternatively, neural networks are used to predict Q-value for each possible action, and then Q-learning can decide which action to take based on the predicted Q-values. Training deep neural networks to approximate the Q function is known as deep Q learning (DQN) \cite{mnih2015human}, which resolves all computational hurdles facing Q-learning. DQN is part of NEORL. Unfortunately, DQN is restricted to discrete spaces.  

This limitation inspires the development of the policy gradient family. PG aims to train a policy that directly maps states to appropriate actions without an explicit Q function in the middle, by optimizing the following loss function

\begin{equation}
\label{eq:pg_loss}
    L^{PG}(\theta) = E_t[log \ \pi_\theta(a_t|s_t)A_t],
\end{equation}
where $E_t$ is the expectation over a batch of state-action pairs, $\pi$ is the policy to be optimized which has weights $\theta$. The policy $\pi$ predicts action $a$ given state $s$ at time step $t$. The term $A_t$ is called the advantage estimate, which expresses the discounted reward $A_t =\sum_{k=0}^\infty \gamma^k r_{t+k}$, where $\gamma$ is the discount factor \cite{schulman2015high}. Most policy gradient algorithms like REINFORCE update the policy parameters through Monte Carlo or random sampling, which introduces high variability in the log of the policy distribution and the reward values (and so $A_t$), leading to unstable learning. This limitation inspired the ``actor-critic'' family with a goal to reduce variance and increase stability through subtracting the discounted reward by a baseline term $V$ as follows
\begin{equation}
\label{eq:adv_est}
    A_t = \underbrace{\sum_{k=0}^\infty \gamma^k r_{t+k}}_\text{Discounted Reward} - \underbrace{V(s_t),}_\text{Baseline (or VF) Estimate of Discounted Reward}
\end{equation}
where $r$ is the reward value, $\gamma$ is the discount factor, and $V$ is the baseline or value function (VF) estimate of the discounted reward. The VF baseline makes the cumulative reward smaller, which makes smaller gradients, and more stable updates. The actor-critic algorithms involve two major phases:
\begin{itemize}
  \item The ``Critic'' phase, which estimates the VF, which could be a Q value.  
  \item The ``Actor'' phase, which updates the policy parameters in the direction suggested by the Critic.
\end{itemize}
Both the critic and the actor functions are parameterized with neural networks, which are updated continuously to maximize the reward return. The NEORL supported algorithms from actor-critic are Advantage Actor Critic (A2C), Actor-Critic with Experience Replay (ACER), Proximal Policy Optimization (PPO), and Actor Critic using Kronecker-Factored Trust Region (ACKTR). See Table \ref{tab:algs} for the full list of the neural-based algorithms within NEORL and their supported spaces. We leveraged our RL optimization classes following the RL implementation provided by OpenAI baselines \cite{baselines} and stable-baselines \cite{stable-baselines}. 

\subsection{Neuroevolution Algorithms}

The category of neuroevolution refers to a class of hybrid algorithms in our framework that utilizes both neural network and evolutionary computation concepts to obtain the best of both worlds. NEORL supports NEAT \cite{stanley2002evolving} for optimization with two different architectures: feedforward neural networks (called FNEAT) and recurrent neural networks (called RNEAT). In both algorithms, GA is used to leverage neural network topology to maximize a fitness function of interest. 

We also provide an implementation for RL-informed EA, a concept developed by the authors \cite{radaideh2021rule}. Using RL as a seeding methodology can be effective to improve the exploration rate of EA in constrained optimization problems. RL algorithms can run as a standalone to generate certain individuals matching part of the problem constraints. Then these RL quality individuals are periodically mixed with EA population to lead to a better search and less local optima entrapment. Currently, in NEORL, the neuroevolution PPO-ES is implemented, where PPO is used to guide genetic algorithms or evolution strategies as illustrated by this large-scale nuclear assembly combinatorial optimization \cite{radaideh2021large}. In addition, we provide an implementation of ACKTR-DE as another neuroevolution alternative that can be promising for continuous optimization. Both of these neuroevolution algorithms feature parallel search.

NEORL provides support for a high-performance hybrid algorithm called PESA (Prioritized replay Evolutionary and Swarm Algorithm), which combines the concept of experience replay with hybrid evolutionary algorithms \cite{radaideh2020improving, schaul2015prioritized}, another concept developed by the authors. Two modes of experience replay are introduced into PESA: prioritized replay to enhance exploration and greedy replay to improve algorithm exploitation. The replay memory stores all individuals in a shared memory across the hybrid algorithms, and replays them with different rates based on their fitness value, such that exploration and exploitation can be balanced during search. Two variants of PESA are currently available in NEORL. Classical PESA, which combines GA, PSO, and SA, and it proved to be effective in discrete optimization. Modern PESA2, which hybridizes GWO, DE, and WOA, proved to be effective in continuous optimization. PESA and PESA2 are candidates to solve expensive problems when computing power is available, as both PESA variants are fully parallelized with parallel efficiency $>85\%$ when tested on 200 processors.  

For expensive optimization with limited computing power, NEORL offers hybrid neuroevolution algorithms, where neural networks act as surrogate models to assist EA search. Currently, we provide an implementation of the recent offline data-driven evolutionary optimization concept \cite{huang2021offline}, where only small number of fitness evaluations are needed to build a surrogate model, such that the surrogate can replace the original fitness function in EA optimization. The original implementation involves using surrogate models of radial basis function networks to assist GA. The authors extend this methodology to more modern algorithms, where we use deep feedforward neural networks as surrogate models through TensorFlow to assist HHO described in section \ref{sec:ea}, where all discrete, continuous, and mixed-discrete spaces are handled. In addition, compared to \cite{huang2021offline}, the neuroevolution Harris hawks optimizer offers parallel training of the surrogate models as well as parallel evaluation of the initial hawks used to construct the surrogate models. These surrogate-based neuroevolution algorithms are useful for large-scale optimization problems with limited computing power. Lastly, Table \ref{tab:algs} shows a summary of all hybrid neuroevolution algorithms in NEORL and their supported spaces.

\subsection{Multiobjective Optimization and Constrained Handling}

Without loss of generality, multiobjective optimization can be formulated as follows: 

\begin{equation}
\label{eq:opt}
\begin{aligned}
    &\underset {\vec{x}}{\text{min } } F(\vec{x})=(f_1(\vec{x}), f_2(\vec{x}), ..., f_k(\vec{x})) \\
    &\text{subject to}, \\
    &g_i(\vec{x}) \geq 0, \quad i= 1, 2, ..., m, \\
    &h_j(\vec{x}) = 0, \quad j= 1, 2, ..., p, \\
    &k \geq 2, \\
\end{aligned}
\end{equation}

where $k$ is the number of objective functions. The set of solutions $\vec{x}$ that satisfies all constraints defines the feasible region $\Omega$, i.e. $\vec{x} \in \Omega$. 

Solving multiobjective optimization can be done either through solving the problem directly by trying to optimize each objective through using posteriori methods, or through using priori methods, which what we currently support in NEROL. Priori methods involve the users by specifying their preferences beforehand to formulate a single-objective optimization problem, such that optimal solutions to the single-objective problem are feasible solutions to the multi-objective problem. Linear scalarization, $\epsilon$-constrained, achievement scalarizing function, and Sen's Multi-Objective Programming are among the common priori methods \cite{cui2017multi}.

The usage of priori methods is advantageous for NEORL, since the formulation of the fitness function will be consistent across EA, neural networks, and neuroevolution. RL algorithms are expected reward (fitness) as a scalar value. Adding special algorithms to NEORL for multiobjective optimization such as non dominated sorting genetic algorithm (NSGA-II) is left for future releases. Currently, to preserve a generic implementation, the user can use variety of priori methods to convert multiobjective to single objective problem. Indeed, NEORL demonstrated successful results with complex multiobjective optimization problems, which are illustrated here \cite{radaideh2021physics, radaideh2021large, radaideh2021rule}, where $\epsilon$-constrained and linear scalarization in conjunction with different neural, evolutionary, and neuroevolution algorithms have been effective in solving the multiobjective optimization.

For constraint handling, we give the user the freedom to handle the constraints internally within the fitness function, before returning the fitness value to the optimizer. This flexibility gives the user a chance to explore different constraint handling methods on the same problem such as applying hard penalty, self-adaptive penalty, or $\epsilon$-constrained \cite{kramer2010review}. 

\section{NEORL Structure}
\label{sec:struct}

In this section, we briefly describe the main steps to do to setup and solve an optimization problem via NEORL. 

\subsection{Fitness/objective Function}
\label{sec:fit}

The fitness function is problem or user-dependent and it is the only NEORL step that depends significantly on the user and application of interest. The fitness function can be as simple as a mathematical function or as complex as a fitness function that has to deal with computer code execution, input handling, and output post-processing. In both cases, the format of the fitness function is similar, it takes input as the vector of input parameters to optimize ($\vec{x}$) and returns output as the scalar fitness value corresponding to this $\vec{x}$. See Python Listing \ref{lst:fit} for a simple fitness function of the multi-dimensional sphere that has the form
\begin{equation}
\label{eq:sphere}
    f(\vec{x}) = \sum_{i=1}^{d} x_i^2,
\end{equation}
where $d$ is the number of dimensions (degrees of freedom) of the Sphere, and it is equal to the size of the input vector $\vec{x}$. For complex fitness functions, the user needs to code more lines to reach the final scalar ``$y$'' value returned by the function. 

\begin{lstlisting}[language=python, label={lst:fit}, basicstyle=\footnotesize, caption={Example of a NEORL fitness function definition}]
def Sphere(x):
    y=sum(xi**2 for xi in x)
    return y
\end{lstlisting}

An additional example on defining advanced fitness function for constrained optimization can be found in Python Listing \ref{lst:constrained} using self-adaptive penalty proposed by \cite{coello2000use}. The single-objective constrained problem is for the three-bar truss design, which is a popular constrained optimization problem described here \cite{mirjalili2015moth}. The problem has three constraints and two input variables to optimize, which are related to the bars' cross-sectional area. Similarly, for multiobjective optimization, the user can change line 24 in Listing \ref{lst:constrained} to a weighted sum of all objectives if linear scalarization is used to convert multiobjective optimization to single objective.  

\begin{lstlisting}[language=python, label={lst:constrained}, basicstyle=\footnotesize, caption={Example of a NEORL fitness function for constrained optimization}]
from math import sqrt
def TBTD(x):
    """
    Three-Bar Truss Design
    See Supplementary Materials Section 2.1 for the Mathematical Description
    """
    x1 = x[0]   #A1=A3 (cross-sectional area)
    x2 = x[1]   #A2 (second cross-sectional area)   
    y = (2*sqrt(2)*x1 + x2) * 100  #objective value (volume of loaded truss structure)

    #Constraints
    g1 = (sqrt(2)*x1+x2)/(sqrt(2)*x1**2 + 2*x1*x2) * 2 - 2  #g1 <= 0 is required 
    g2 = x2/(sqrt(2)*x1**2 + 2*x1*x2) * 2 - 2               #g2 <= 0 is required
    g3 = 1/(x1 + sqrt(2)*x2) * 2 - 2                        #g3 <= 0 is required
    g = [g1,g2,g3]
    
    #penalty coefficients
    w1=100   #coefficient for self-adaptive penality (Coello, 2000) 
    w2=100   #coefficient for self-adaptive penality (Coello, 2000) 
    
    #final fitness value
    phi=sum(max(item,0) for item in g)     #sum of all constraint values
    viol=sum(float(num) > 0 for num in g)  #sum of the number of violated constraints
    fitness=y + w1*phi + w2*viol           #penalized fitness value
    
    return fitness
\end{lstlisting}

\subsection{Input Search Space}
\label{sec:space}

NEORL supports three types of input spaces: discrete variable (labelled by \code{int}), continuous variable (labelled by \code{float}), and categorical variable (labelled by \code{grid}). A mixed space of the three types is also possible to define. A discrete variable is expected to take integer values, where the label \code{int} is followed by the lower and upper bounds allowed. The continuous variable takes real values,  where the label \code{float} is followed by the lower and upper bounds allowed. The categorical variable takes fixed values defined by a grid, where the label \code{grid} is followed by the list of values as a tuple (i.e. enclosed in a parenthesis), e.g. (value 1, value 2, ...). The values in the grid can be integers, real, and/or of string type. NEORL expects the full parameter space to be defined in Python dictionary (\code{dict}) format. See Listing \ref{lst:space} which demonstrates two examples of how to define an input space $\vec{x}$ with NEORL. In addition, all possibilities of input spaces are defined in Table \ref{tab:space} with a demonstration example. The reader can check the space types that each NEORL algorithm supports in Table \ref{tab:algs}.

\begin{lstlisting}[language=python, basicstyle=\footnotesize, label={lst:space}, caption={Example of a NEORL search space definition}]
#A real-valued input space of size 5
d=5
Space1={}
for i in range(1,d+1):
    Space1['x'+str(i)]=['float', -100, 100]  #5 continuous variables between -100/100
#A mixed input space of size 4
Space2 = {}
Space2['n_layers'] = ['int', 1, 6]  #number of layers in a network as integer
Space2['n_nodes'] = ['int', 50, 150]  #number of nodes in each layer as integer
Space2['activation'] = ['grid', ('linear', 'rbf', 'sigmoid')]  #activation type as categorical 
Space2['learning_rate'] = ['float', 1e-4, 1e-3]  #learning rate as continuous variable
\end{lstlisting}

% Table generated by Excel2LaTeX from sheet 'spaces'
\begin{table}[htbp]
  \centering
  \caption{Input space types supported by NEORL}
    \begin{tabular}{rl}
    \toprule
    \multicolumn{1}{l}{Space Type} & Example \\
    \midrule
    \multicolumn{1}{l}{Combinatorial/Discrete} & space[\textquotesingle x1\textquotesingle] = [\textquotesingle int\textquotesingle, 1, 6]  \\
          & space[\textquotesingle x2\textquotesingle] = [\textquotesingle int\textquotesingle, 50, 150] \\
          & space[\textquotesingle x3\textquotesingle] = [\textquotesingle int\textquotesingle, -10, 10] \\
    \midrule
    \multicolumn{1}{l}{Categorical} & space[\textquotesingle x1\textquotesingle] = [\textquotesingle grid\textquotesingle, (1, 2, 3, 4, 5, 6)]  \\
          & space[\textquotesingle x2\textquotesingle] = [\textquotesingle grid\textquotesingle, (\textquotesingle linear\textquotesingle, \textquotesingle rbf\textquotesingle, \textquotesingle sigmoid\textquotesingle)] \\
          & space[\textquotesingle x3\textquotesingle] = [\textquotesingle grid\textquotesingle, (1.5, 2.5, 3.5, 4.5)] \\
    \midrule
    \multicolumn{1}{l}{Continuous} & space[\textquotesingle x1\textquotesingle] = [\textquotesingle float\textquotesingle, 1, 6]  \\
          & space[\textquotesingle x2\textquotesingle] = [\textquotesingle float\textquotesingle, 50, 150] \\
          & space[\textquotesingle x3\textquotesingle] = [\textquotesingle float\textquotesingle, -10, 10] \\
    \midrule
    \multicolumn{1}{l}{Mixed Discrete/Categorical} & space[\textquotesingle x1\textquotesingle] = [\textquotesingle int\textquotesingle, 1, 6]  \\
          & space[\textquotesingle x2\textquotesingle] = [\textquotesingle grid\textquotesingle, (\textquotesingle linear\textquotesingle, \textquotesingle rbf\textquotesingle, \textquotesingle sigmoid\textquotesingle)] \\
          & space[\textquotesingle x3\textquotesingle] = [\textquotesingle grid\textquotesingle, (1, 2.5, 5, 10)] \\
    \midrule
    \multicolumn{1}{l}{Mixed Discrete/Continuous} & space[\textquotesingle x1\textquotesingle] = [\textquotesingle int\textquotesingle, 1, 6]  \\
          & space[\textquotesingle x2\textquotesingle] = [\textquotesingle float\textquotesingle, 50, 150] \\
          & space[\textquotesingle x3\textquotesingle] = [\textquotesingle float\textquotesingle, -10, 10] \\
    \midrule
    \multicolumn{1}{l}{Mixed Categorical/Continuous} & space[\textquotesingle x1\textquotesingle] = [\textquotesingle float\textquotesingle, -100, 100]  \\
          & space[\textquotesingle x2\textquotesingle] = [\textquotesingle grid\textquotesingle, (\textquotesingle linear\textquotesingle, \textquotesingle rbf\textquotesingle, \textquotesingle sigmoid\textquotesingle)] \\
          & space[\textquotesingle x3\textquotesingle] = [\textquotesingle grid\textquotesingle, (1, 2.5, 5, 10)] \\
    \midrule
    \multicolumn{1}{l}{Mixed Discrete/Continuous/Categorical} & space[\textquotesingle x1\textquotesingle] = [\textquotesingle int\textquotesingle, 1, 6]  \\
          & space[\textquotesingle x2\textquotesingle] = [\textquotesingle grid\textquotesingle, (\textquotesingle linear\textquotesingle, \textquotesingle rbf\textquotesingle, \textquotesingle sigmoid\textquotesingle)] \\
          & space[\textquotesingle x3\textquotesingle] = [\textquotesingle float\textquotesingle, -10, 10] \\
    \bottomrule
    \end{tabular}%
  \label{tab:space}%
\end{table}%

\subsection{Algorithms}

Every algorithm class in NEORL is expecting multiple arguments to establish an algorithm instance, which can be grouped into: (1) problem-based, (2) algorithm hyperparameters, and (3) miscellaneous. The problem-based arguments are the \code{mode} of the problem (minimization or maximization), the \code{bounds} which is the parameter space defined in section \ref{sec:space}, and the fitness function (\code{fit}) defined in section \ref{sec:fit}. Algorithm hyperparameters are specific to each algorithm, which usually need to be tuned by the user for optimal performance. Miscellaneous arguments may include the random \code{seed} for reproducibility and \code{ncores} for multiprocessing (see section \ref{sec:parallel}). After defining an algorithm instance, the user can start the optimization process by a special method called \code{.evolute} for evolutionary algorithms and \code{.learn} for neural and RL algorithms. 

For EA, these methods expect number of generations/iterations to run the optimizer for (e.g. \code{ngen}), an initial guess \code{x0} if applicable (if \code{x0=None} the optimizer starts with a random guess), and \code{verbose} to print the progress. See Listing \ref{lst:algs}, which shows two examples of setting and evolving a grey wolf and a simulated annealing optimizer. The listing shows that number of wolves in the group (\code{nwolves}) and size of the annealing chain before update (\code{chain\_size}) are a few examples of the hyperparameters of these algorithms. At the end of evolution, the optimizer returns three variables: \code{xbest}, \code{ybest}, and \code{log}, which are respectively: (1) the best value of the input vector $\vec{x}$, the best value of the fitness $y$, and a logger in dictionary form containing major statistics as found by the optimizer.  

\begin{lstlisting}[language=python, basicstyle=\footnotesize, label={lst:algs}, caption={Example of NEORL evolutionary algorithm setup}]
from neorl import GWO, SA    #import the classes
#Grey Wolf Optimization (setup and evolution)
gwo=GWO(mode='min', fit=Sphere, bounds=Space1, nwolves=5, <@\textcolor{red}{ncores=1}@>, seed=1)
xbest, ybest, log=gwo.evolute(ngen=100, x0=None, verbose=True)

#Simulated Annealing (setup and evolution)
sa=SA(mode='min', fit=Sphere, bounds=Space1, chain_size=30, chi=0.2, <@\textcolor{red}{ncores=1}@>, seed=1)
xbest, ybest, log=sa.evolute(ngen=100, x0=None, verbose=True)
\end{lstlisting}

For neural algorithms, these methods expect definition of special classes to be compatible with the RL paradigm (action, state, reward, etc.). Listing \ref{lst:algs2} shows how to set and evolve a PPO optimizer. First, a RL environment is constructed by passing some problem-based arguments such as the fitness function (\code{fit}), the bounds of the input space (\code{bounds}), problem mode (\code{mode}), number of processors (\code{ncores}), and the length of the episode in time steps (\code{episode\_length}). Afterwards, the user defines a logger using the method \code{RLLogger} which records samples during training with frequency \code{check\_freq}. The PPO object can then be constructed by defining the policy type such as feedforward policy (\code{MlpPolicy}), passing the created environment instance (\code{env}), and a set of PPO hyperparameters (e.g. \code{n\_steps}, \code{ent\_coef}). The learning process can be started using the \code{.learn} method. The user should notice here that neural algorithms \code{learn} function expects a total of fitness evaluations in time steps (\code{total\_timesteps}). The user can calculate the number of generations as in EA by simply taking the ratio of \code{total\_timesteps} over \code{episode\_length}. Finally, the user can access the best solutions from the callback function via the variables: \code{cb.xbest} and \code{cb.rbest}.

\begin{lstlisting}[language=python, basicstyle=\footnotesize, label={lst:algs2}, caption={Example of NEORL neural algorithm setup}]
from neorl import PPO2, MlpPolicy, RLLogger, CreateEnvironment

#For neural algorithms, create an environment that includes the fitness, space type, and problem mode (min/max)
env=CreateEnvironment(method='ppo', fit=Sphere, bounds=bounds, mode='min', <@\textcolor{red}{ncores=1}@>, episode_length=50)
cb=RLLogger(check_freq=1, save_model=True)                              #create a callback function to log data every step
#MlpPolicy is a feedforward multilayer perceptron neural network policy
ppo = PPO2(MlpPolicy, env=env, n_steps=12, ent_coef=0.01, seed=1)     #create a PPO object based on the defined env
#number of generations=total_timesteps/episode_length
ppo.learn(total_timesteps=2000, callback=cb)           #optimize the environment fitness
#access optimal results
print('The best value of x found:', cb.xbest)
print('The best value of fitness found:', cb.rbest)
\end{lstlisting}

\subsection{Parallel Computing}
\label{sec:parallel}

Parallel optimization is crucial to accelerate the process if the fitness function is very expensive to evaluate (e.g. requires running a computer code). Fortunately, invoking parallel optimization via NEORL is very straightforward via the argument \code{ncores}, where the user specifies number of processors to use, see Listing \ref{lst:algs}-\ref{lst:algs2}. Currently, majority of NEORL algorithms support parallel/multiprocessing optimization (see Table \ref{tab:algs}). 

\begin{figure}[h] 
 \centering
  \includegraphics[width=0.5\textwidth]{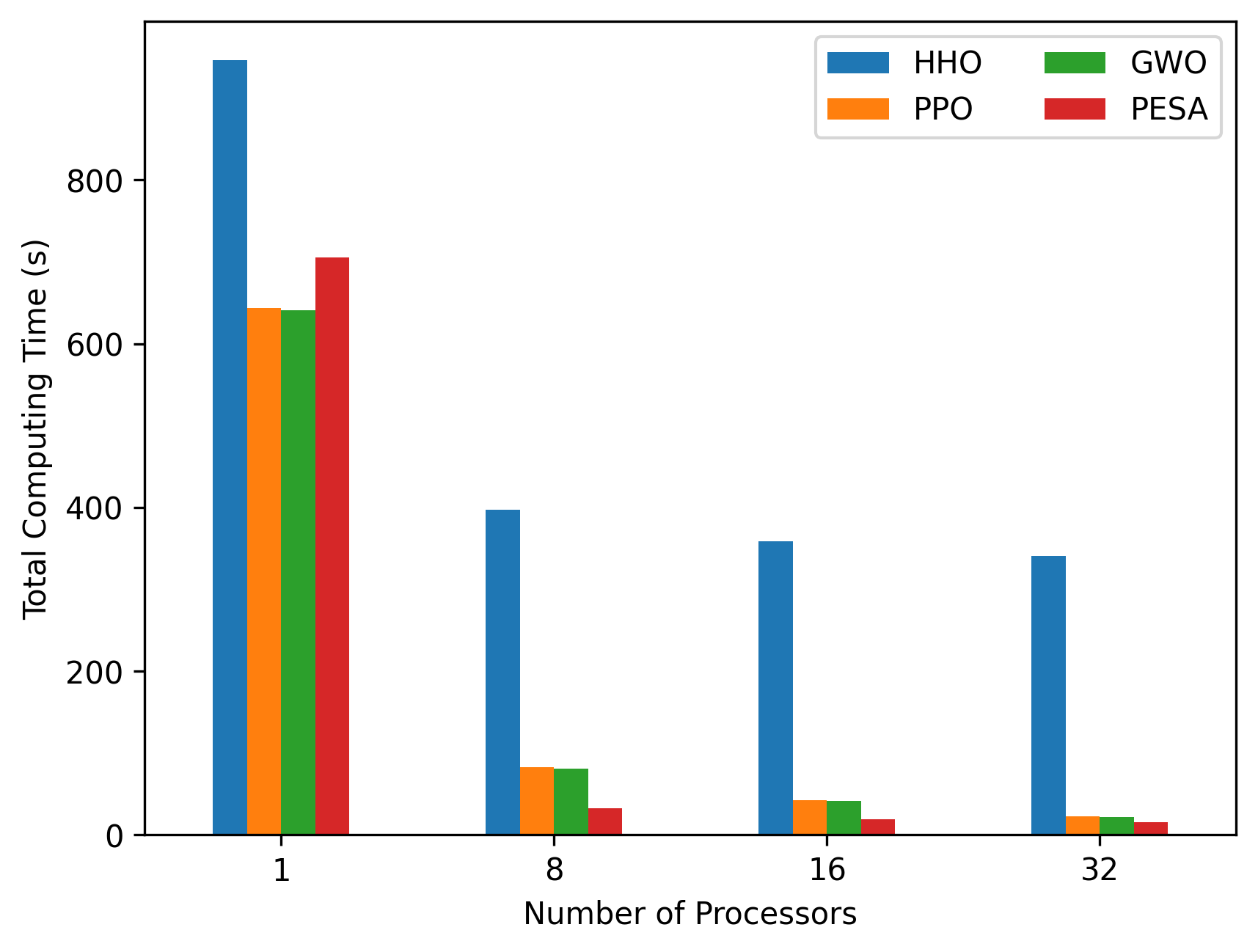}
  \caption{Comparison of multiprocessing performance of different NEORL algorithms}
  \label{fig:parallel}
\end{figure}

Figure \ref{fig:parallel} compares parallel optimization performance of four different algorithms from the three different categories in optimizing the sphere function. The sphere function is delayed by 1s to reflect expensive optimization and to avoid conflict with parallel overhead costs. All algorithms have been executed for 20 generations and 32 individuals per generation using 1, 8, 16, and 32 processors. First we can notice a large decrease in computational time when increasing the number of processors for all four algorithms. Second, we can notice that HHO is significantly slower than other algorithms, even compared to its closest EA fellow, GWO. The algorithm steps of GWO can be completely parallelized due to the nature of the algorithm. However, HHO algorithm has some serial steps, in particular, when executing soft and hard besieges, which require dependency between the population individuals. This makes HHO multiprocessing slower than GWO. Third, we can see that PESA is slower than PPO and GWO when using single processor, while its superior performance is clearly observed after activating its two-levels of parallelism. Overall, in Figure \ref{fig:parallel}, moving from 1 processor to 32, the speedup factor of HHO is only 3, GWO is 30, PPO is also 30, and PESA is 46.   

\subsection{Restart Capabilities}

NEORL provides restart capabilities for the user to restart the search from a recent checkpoint. This feature is useful for long optimization runs, when the search could terminate due to technical issues or power outage. For the RL neural optimizers, the neural network model can be saved periodically using the flag \code{save\_model} in the \code{RLLogger}, see line 5 of Listing \ref{lst:algs2}. Similarly, for the evolutionary or neuroevolution optimizers, the user can use the argument \code{x0} (see line 4 or line 8 of Listing \ref{lst:algs}), to specify the initial population. The user can recover a population using the best individuals of the previous generations, and use that population to continue the search.    

\subsection{Hyperparameter Tuning}
\label{sec:hyper}

In certain cases when the problem is high-dimensional and the fitness function is very expensive and complex, the performance of the algorithms may become sensitive to their hyperparameters (e.g. \code{nwolves}, \code{chi}, \code{chain\_size}, \code{n\_steps}). To keep the user isolated from the tedious burden of manual and trial-and-error tuning, NEORL provides four systematic and automatic methods to assess the optimizer performance under different combination of hyperparameters for a certain problem/fitness setup. The top hyperparameter sets are reported by these methods for the user to use for further analysis. The tuning methods are: exhaustive grid search, random search, Bayesian search, and evolutionary search. In addition, most of these methods support parallel search to significantly accelerate the tuning process. We refer the user to this link\footnote{Grid Search: \url{https://neorl.readthedocs.io/en/latest/tune/grid.html}}, which contains a Python example of how to use NEORL grid search, and other methods can be accessed from the same webpage. 

\begin{figure}[h] 
 \centering
  \includegraphics[width=0.5\textwidth]{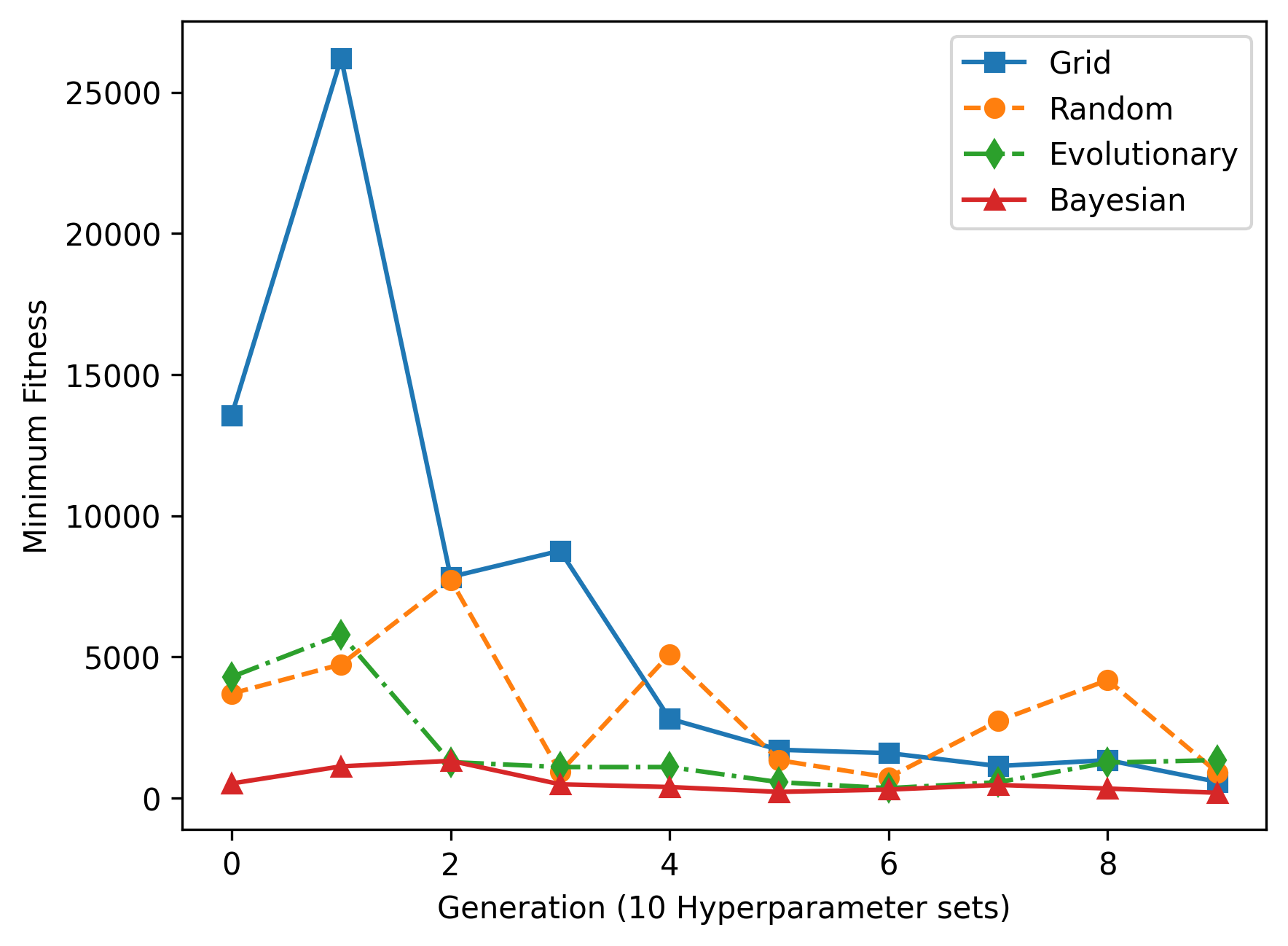}
  \caption{Comparison of the hyperparameter tuning algorithms in NEORL}
  \label{fig:hyperparam}
\end{figure}

We compare the hyperparameter search algorithms in tuning ES/GA to minimize a 20-dimension sphere function of Eq.\eqref{eq:sphere}. We target three hyperparameters of ES: \code{cxpb}, \code{mutpb}, and \code{alpha}, which are respectively the crossover probability, mutation probability, and probability of blending during crossover (i.e. all hyperparameters have a range of 0 to 1). All four methods have been executed to evaluate 100 combinations of the previous hyperparameters, and the minimum fitness of every 10 combinations is plotted in Figure \ref{fig:hyperparam} for consistency across methods. All methods seem to perform in a comparable manner given the converged fitness is close between the four methods. Bayesian and evolutionary methods seem to show quicker convergence than grid and random searches due to their systematic search behaviour. Bayesian search found the best hyperparameter set as follows: \code{cxpb = 0.8}, \code{mutpb = 0.2}, and \code{alpha = 0.424}.

Lastly, we end this section by Listing \ref{lst:ex}, which shows how to setup a full optimization problem in NEORL with few lines. The problem in Listing \ref{lst:ex} shows how to optimize the 5-d Sphere function of Eq.\eqref{eq:sphere} with the grey wolf optimizer (GWO). To further validate NEORL capabilities, the framework is benchmarked against a variety of mathematical functions, which are not reported here for brevity and kept for the full paper.

\begin{lstlisting}[language=python, basicstyle=\footnotesize, label={lst:ex}, caption={NEORL full example of optimizing the 5-d Sphere}]
from neorl import GWO
#Setup the fitness function
def Sphere(x) :
    y=sum(xi**2 for xi in x)
    return y
#Setup the parameter space
d=5
Space={}
for i in range (1, d+1):
    Space['x'+str(i)]=['float', -100, 100]
#Setup and evolute the optimizer
gwo=GWO(mode='min', fit=Sphere, bounds=Space, nwolves=5, ncores=1, seed=1)
xbest, ybest, log=gwo.evolute(ngen=100, verbose=1)
print('Best input: ', xbest) 
print('Best fitness:', ybest)
#which prints these after running:
#Best input:  [0.00015237, -0.0002251, 0.00023039, -0.00016578, 0.00019899]
#Best fitness: 1.9404338787767446e-07
\end{lstlisting}

\subsection{Comparison with Other Frameworks}

Table \ref{tab:pycomp} compares NEORL to the popular Python optimization frameworks based upon different features that each user prioritizes. The comprehensive nature of NEORL is clearly displayed. For gradient-based algorithms, they are not explicitly implemented in NEORL due to the fact that these algorithms are well-supported in other frameworks such as PyOpt \cite{perez2012pyopt} and nlopt \cite{johnson2014nlopt}. NEORL provides access to more algorithms, diverse algorithm types, hyperparameter tuning, parallel computing, detailed documentation, and real-world engineering examples. Here we can highlight that ``PyOpt'' \cite{perez2012pyopt} is recommended for gradient-based algorithms, even though the framework is not continuously maintained or updated by the authors. Also, ``PyOpt'' is only restricted to the legacy Python 2, which is superseded by the modern Python 3. ``nlopt'' \cite{johnson2014nlopt} provides robust gradient-based and evolutionary algorithms, however, without parallel support or detailed documentation of the Python plugin. ``EvoloPy'' \cite{faris2016evolopy} provides a good Python implementation of modern evolutionary algorithms (e.g. GWO, WOA, SSA), while the framework seems to lack detailed documentation, parallel computing, and tests on large-scale problems. Lastly, ``DEAP'' \cite{DEAP_JMLR2012} is undoubtedly the most popular evolutionary optimization framework when it comes to genetic algorithms and evolution strategies, with a detailed documentation and real-world examples. Nevertheless, the lack of support of modern EA, neural algorithms, and hybrid neuroevolution can be considered as an advantage for NEORL over DEAP.

\begin{table}[htbp]
  \centering
  \small
  \caption{Comparison of NEORL with other popular Python optimization frameworks}
  \begin{threeparttable}
    \begin{tabular}{llllll}
    \toprule
    Feature & NEORL & PyOpt \cite{perez2012pyopt} & EvoloPy \cite{faris2016evolopy} & nlopt \cite{johnson2014nlopt} & DEAP \cite{DEAP_JMLR2012} \\
    \midrule
    Number of Algorithms & 28    & 20    & 14    & 23    & 5 \\
    Classical Evolutionary Computation & \cmark & \cmark & \cmark & \cmark & \cmark \\
    Modern Evolutionary Computation & \cmark & \xmark & \cmark & \xmark & \xmark \\
    Neural Networks & \cmark & \xmark & \xmark & \xmark & \xmark \\
    Gradient-based Algorithms & \cmark* & \cmark & \xmark & \cmark & \xmark \\
    Hybrid Algorithms & \cmark & \xmark & \xmark & \xmark & \xmark \\
    Hyperparameter Tuning & \cmark & \xmark & \xmark & \xmark & \xmark \\
    Parallel Computing & \cmark & \cmark & \xmark & \xmark & \cmark \\
    Detailed Documentation & \cmark & \cmark & \xmark & \xmark**  & \cmark \\
    Real-world Examples  & \cmark & \cmark & \xmark & \xmark** & \cmark \\
    \bottomrule
    \end{tabular}%
    \begin{tablenotes}
      \small
      \item $*$ Gradient-based algorithms in NEORL are embedded as optimizers of the neural networks  
      \item $**$ \code{nlopt} is implemented in different programming languages, the Python plugin is not thoroughly documented/benchmarked. 
    \end{tablenotes}
    \end{threeparttable}
  \label{tab:pycomp}%
\end{table}%

\section{Applications}
\label{sec:apps}

In this section, we apply NEORL to two different carbon-free energy applications on fuel cell power production (continuous optimization) and nuclear reactor control (mixed discrete-continuous optimization) to demonstrate NEORL potential to help combating climate change. Links for additional NEORL applications are listed by the end of this section. 

\subsection{Fuel Cell Design Optimization}
\label{sec:fc}

A solid oxide fuel cell (SOFC) converts chemical energy into electricity via the oxidation of a fuel, typically hydrogen \cite{ni2007parametric, zhang2017two}. The topic of fuel cell design optimization is continuously researched to improve the power output and efficiency using variety of optimization, uncertainty analysis, and machine learning methods \cite{inci2020performance, kannan2020quantifying, mojaver2021combined, radaideh2020design}. Here we seek a similar objective to optimize various operating parameters of a SOFC with the goal of maximizing both the power output, $P$, and the efficiency, $\eta$, consolidated into a single objective function given by
\begin{equation}
  \max_{\vec{x}} f(\vec{x}) = 0.5 \times P + 0.5 \times \eta \times 100.
\end{equation}

The voltage in the fuel cell is calculated as 
\begin{equation}
\label{eq:V_cell}
    V_{cell} = E-V_{act}-V_{ohm}-V_{conc},
\end{equation}
where $E$ is the maximum voltage, $V_{act}$ is the activation overpotential due to electrode kinetics, $V_{ohm}$ is the ohmic overpotenial associated with electrical resistance, and $V_{conc}$ is the concentration overpotential caused by resistance to the movement of reactants and products \cite{radaideh2020efficient}.

The maximum voltage is expressed as
\begin{equation}
    E = \frac{-\Delta g}{n_eF} + \frac{RT}{n_eF}ln\left ( \frac{p_{H_2}p_{O_2}^{0.5}}{p_{H_2O}} \right ),
\end{equation}
where $\Delta g = -282150 + 86.735 T$ is the Gibbs free energy change, $n_e$ is the number of electrons per reaction, $F$ is Faraday's constant, $R$ is the universal gas constant, $T$ is the operating temperature of SOFC, and $p_{H_2}$, $p_{O_2}$, and $p_{H_2O}$ are the partial pressure of hydrogen (H$_2$), oxygen (O$_2$), and water (H$_2$O), respectively \cite{radaideh2020efficient}.

The total activation overpotential is the sum of both the anode and cathode parts, $V_{act,a}$ and $V_{act,c}$. The equation for $V_{act,k}$ for $k=a,c$ is \cite{ni2007parametric,zhang2017two,zhang2013performance}
\begin{equation}
\begin{aligned}
   V_{act,k} &= \frac{RT}{F}sinh^{-1}\left(\frac{j}{2j_{0,k}}\right) \\ &= \frac{RT}{F} ln\left [\frac{j}{2j_{0,k}}+ \sqrt{\left ( \frac{j}{2j_{0,k}} \right )^2+1}\right] (k = a,c), % second line  
\end{aligned}
\end{equation}
where $j$ is the operating current density and $j_{0,k}$ is the exchange current density at either the anode or the cathode. The exchange current densities at the anode and cathode are given respectively as \cite{ni2007parametric,zhang2017two,zhang2013performance}
\begin{equation}
j_{0,a}=\gamma_a\frac{72X[D_p-(D_p+D_s)\epsilon]\epsilon}{D_{s}^2D_{p}^2(1-\sqrt{1-X^2})} \times \left(\frac{p_{H_2}}{p}\right)\left(\frac{p_{H_2O}}{p}\right) exp\left ( -\frac{E_{act,a}}{RT} \right ),   
\end{equation}
and
\begin{equation}
j_{0,c}=\gamma_c\frac{72X[D_p-(D_p+D_s)\epsilon]\epsilon}{D_{s}^2D_{p}^2(1-\sqrt{1-X^2})} \times \left(\frac{p_{O_2}}{p}\right)^{0.25}exp\left (-\frac{E_{act,c}}{RT} \right ),    
\end{equation}
where $\gamma_a$ and $\gamma_c$ are the coefficients of exchange current density at the anode and cathode, $X$ is the ratio of grain contact neck length to the grain size, $D_p$ and $D_s$ are pore size and grain size, respectively, $\epsilon$ is the electrode porosity, $p$ is the operating pressure, and $E_{act,a}$ and $E_{act,c}$ are the activation energy for the anode and cathode, respectively.

The ohmic overpotential is given by \cite{ni2007parametric}
\begin{equation}
    V_{ohm}=j\left (\frac{L_a}{\sigma_a}+\frac{L_c}{\sigma_c}+\frac{L_e}{\sigma_e} \right),
\end{equation}
where $L_a$, $L_c$, and $L_e$ are the thicknesses of the anode, cathode, and electrolyte, respectively and $\sigma_a$ and $\sigma_c$ are the electric conductivity of the anode and cathode, respectively. The ionic conductivity of the electrolyte, $\sigma_e$, is
\begin{equation}
    \sigma_e=exp(-1.03 \times 10^4/T).
\end{equation} 

Finally, the concentration overpotential can be expressed as the sum of the anode and cathode parts, $V_{conc,a}$ and $V_{conc,c}.$ At the anode, the concentration overpotential is
\begin{equation}
    V_{conc,a}= \frac{RT}{n_eF}\left [ ln\left(1+\frac{j}{j_{lH_2O}}\right)- ln \left(1-\frac{j}{j_{lH_2}}\right) \right ],
\end{equation}
where $j_{lH_2O} = 2.27 \times 10^7$ A m$^{-2}$ and $j_{lH_2} = 4 \times 10^5$ A m$^{-2}$ \cite{radaideh2020efficient}.

At the cathode, the concentration overpotential is \cite{ni2007parametric,zhang2017two,zhang2013performance}
\begin{equation}
V_{conc,c}=\frac{RT}{4F}ln\left[\frac{p_{O_2}}{\frac{p_c}{D}-\left ( \frac{p_c}{D}-p_{O_2}\right)exp\left(\frac{RTL_cjD}{4FD^{eff}_{O_2}p_c} \right )}\right],
\end{equation}
where $p_c = p_{O_2} + p_{N_2}$ is the operating pressure on the cathode, $D^{eff}_{O_2}$ is the effective diffusion coefficient of the O$_2$ species, and $D$ is the ratio of diffusion coefficients, $D=D^{eff}_{O_2,Kn}/(D^{eff}_{O_2,Kn}+D^{eff}_{O_2-N_2})$.

All of these equations are combined back into Eq.\eqref{eq:V_cell}, so the power output and efficiency can be calculated respectively as
\begin{equation}
    P=jAV_{cell}
\end{equation}
and
\begin{equation}
    \eta = \frac{P}{-\Delta\dot{H}}=-\frac{n_eFV_{cell}}{\Delta h},
\end{equation}
where $\Delta \dot{H}= (\Delta h /n_e F) j A$ is the total energy content of the SOFC fuel per unit time.

In total there are 20 uncertain parameters to be optimized with lower and upper bounds shown in Table \ref{tab:SOFC_table}. The bounds are derived from the distributions outlined by \cite{radaideh2020efficient}, where parameters with uniform distributions are optimized over their entire domain, while parameters with normal distributions are bounded within four standard deviations of the mean. In this analysis, the operating temperature $T$ is held constant.

\begin{table}[h]
\centering
\small
\caption{\label{tab:SOFC_table}The upper and lower bounds of all 20 uncertain parameters in the fuel cell}
\begin{tabular}{cccccccc}
\toprule
Parameter & Bounds\\
\hline
Current density, $j^*$ (A m$^{-2}$) & $[12672, 13728]$ \\
Effective surface area, $A$ (m$^2$) & $[1.55 \times 10^{-3}, 1.65 \times 10^{-3}]$ \\
Partial pressure of O$_2$, $p_{O_2}$ (atm) & $[0.168, 0.252]$ \\
Partial pressure of H$_2$O, $p_{H_2O}$ (atm) & $[0.04, 0.06]$  \\
Electrode porosity, $\epsilon$  & $[0.3936, 0.5664]$ \\
Electrode tortuosity, $\xi$  & $[4.32, 6.48]$ \\
Average pore diameter, $D_p$ (m)  & $[2.8 \times 10^{-6}, 3.2 \times 10^{-6}]$ \\
Average grain size, $D_s$ (m) & $[1.4 \times 10^{-6}, 1.6 \times 10^{-6}]$ \\
Average length of grain contact, $X$ & $[0.6,0.8]$ \\
Anode thickness, $L_a$ (m)  & $[4.7 \times 10^{-4}, 5.3 \times 10^{-4}]$ \\
Cathode thickness, $L_c$ (m) & $[4.7 \times 10^{-5}, 5.3 \times 10^{-5}]$ \\
Electrolyte thickness, $L_e$ (m) & $[4.7 \times 10^{-5}, 5.3 \times 10^{-5}]$ \\
Anode electric conductivity, $\sigma_a$ ($\Omega^{-1}$ m) & $[48000, 112000]$ \\
Cathode electric conductivity, $\sigma_c$ ($\Omega^{-1}$ m) & $[5040, 11760]$ \\
Diameters of the H$_2$ molecular collision,  $\sigma_{H_2}$ ($\AA$) & $[2.159, 3.495]$ \\
Diameters of the H$_2$O molecular collision,  $\sigma_{H_2O}$ ($\AA$) & $[2.009, 3.273]$ \\
Diameters of the O$_2$ molecular collision,  $\sigma_{O_2}$ ($\AA$) & $[2.635, 4.299]$ \\
Diameters of the N$_2$ molecular collision,  $\sigma_{N_2}$ ($\AA$) & $[2.886, 4.710]$ \\
Anode coefficient of the exchange current density, $\gamma_a$ (A m) & $[1.39 \times 10^{-9}, 1.69 \times 10^{-9}]$ \\
Cathode coefficient of the exchange current density, $\gamma_c$ (A m) & $[5.27 \times 10^{-10}, 6.44 \times 10^{-10}]$ \\
\bottomrule
\end{tabular}
\end{table}

\begin{figure}[!h]
    \centering
    \includegraphics[width=0.45\textwidth]{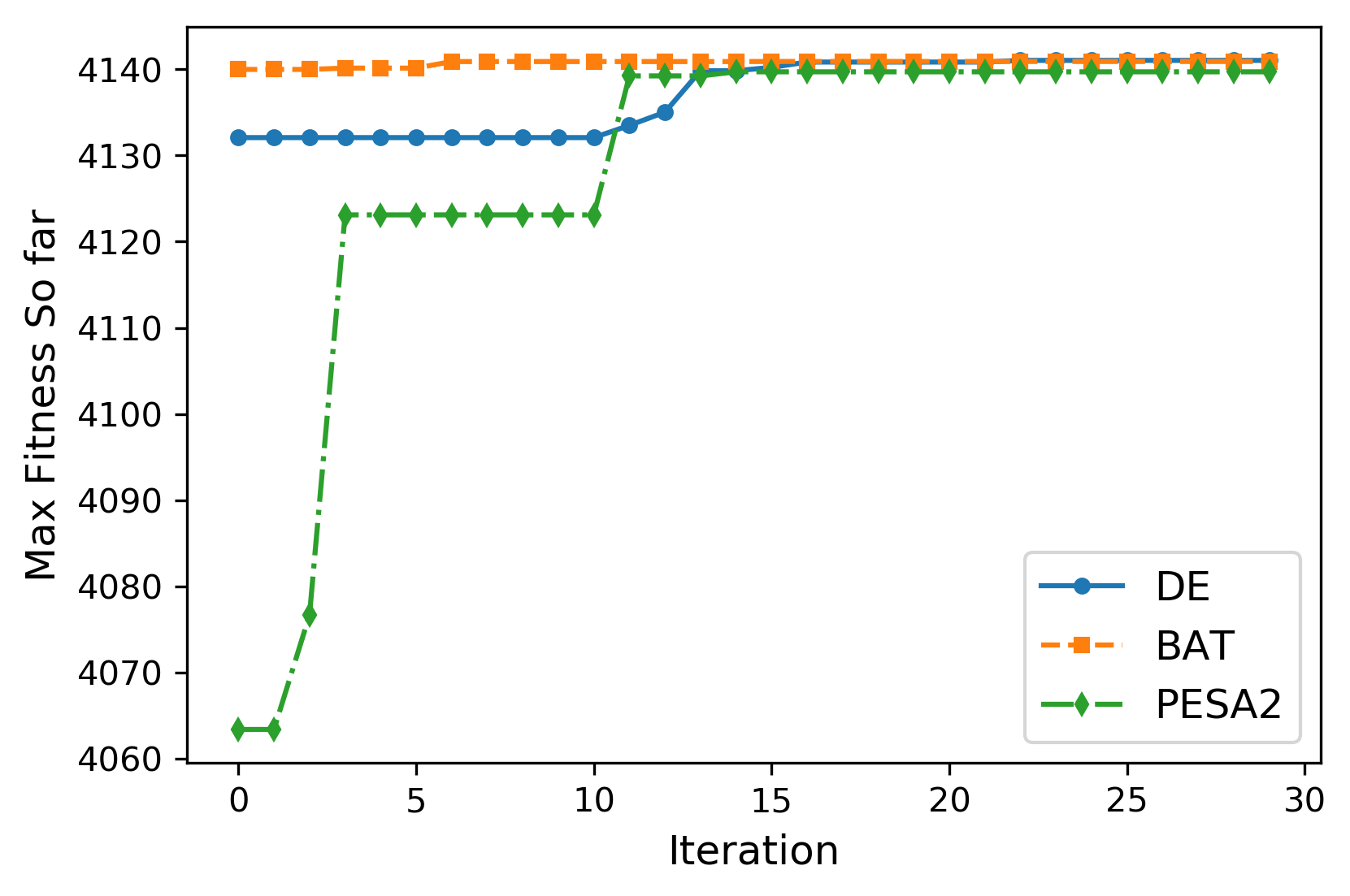}
    \caption{The convergence of the fuel cell fitness function using Bayesian hyperparameter tuning on DE, BAT, and PESA2 (GWO and MFO are hyperparameter-free).}
    \label{fig:bayesian}
\end{figure}

This problem is solved using differential evolution (DE), bat algorithm (BAT), grey wolf optimizer (GWO), moth-flame optimization (MFO), and modern PESA (PESA2) within NEORL. Each algorithm is executed for 300 generations with 50 individuals for a total of 15000 fitness evaluations. In addition, Bayesian search is applied to 3 algorithms: BAT, PESA2, and DE for 30 iterations to tune their hyperparameters with reduced number of generations per iteration. GWO and MFO are not tuned as they are considered hyperparameter-free, since we fixed the population size between the five algorithms. Figure \ref{fig:bayesian} shows the convergence of the Bayesian hyperparameter tuner for DE, BAT, and PESA2. We can notice that the Bayesian search results indicate finding an optimal hyperparameter set within 15 iterations, which is very efficient. Also, the results show large sensitivity of PESA2 to its hyperparameters (which is expected for hybrid algorithms), followed by DE. However, we can see that BAT has small sensitivity, as BAT fitness slightly improved during hyperparameter search.

\begin{figure}[!h]
    \centering
    \includegraphics[width=0.45\textwidth]{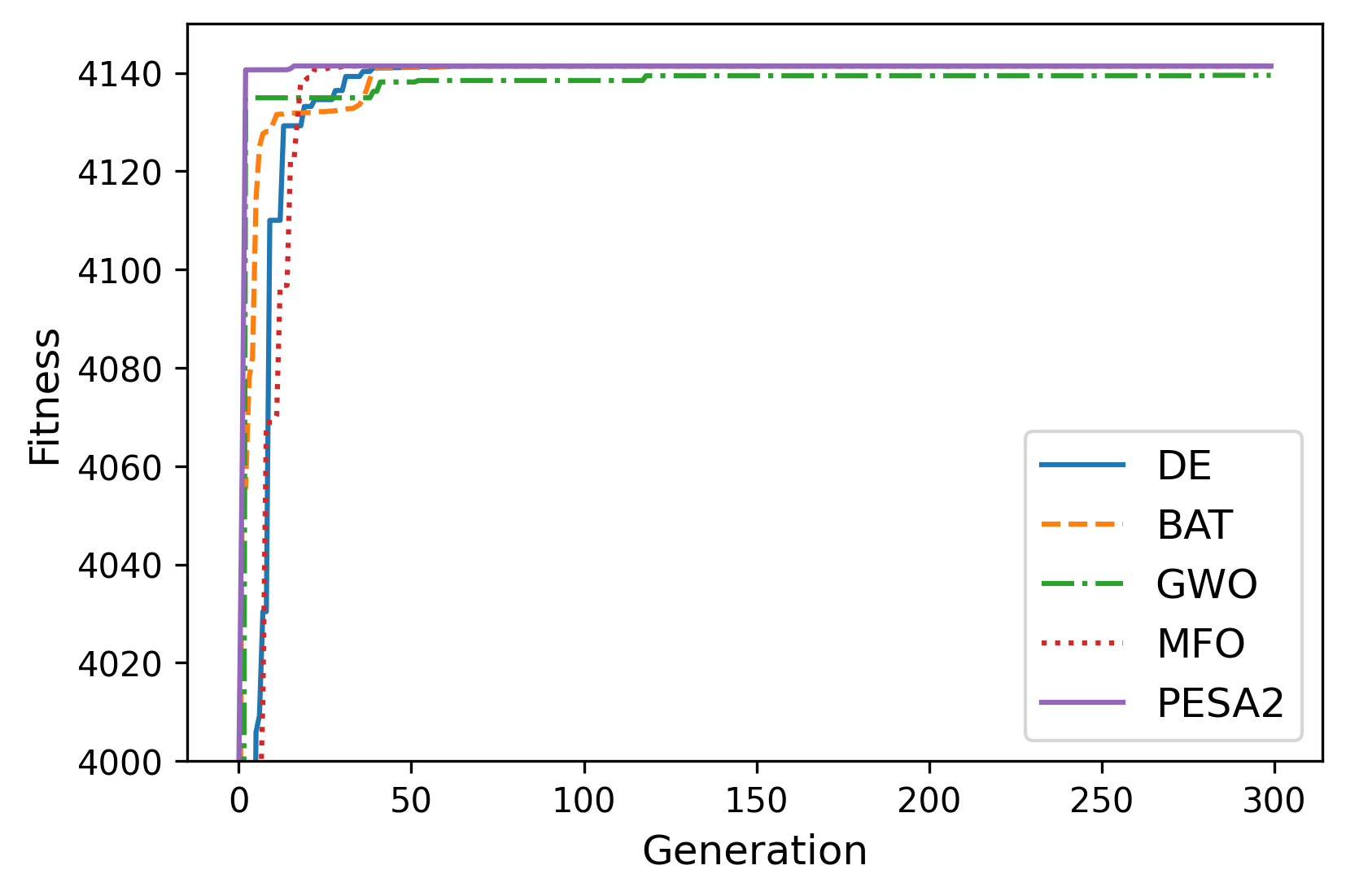}
    \caption{The convergence of the fuel cell fitness function using DE, BAT, GWO, MFO, and PESA2}
    \label{fig:SOFC_convergence}
\end{figure}

The convergence of the fuel cell fitness function for the five algorithms under their optimal hyperparameters is shown in Figure \ref{fig:SOFC_convergence} after running for 300 generations. Identical maximum solutions are found by DE, BAT, MFO, and PESA2 with a fitness value of $4141.4011$, corresponding to $P=8236.16$ W and $\eta=46.64\%$. Although most algorithms converge to similar solutions, PESA2 and MFO show faster convergence than other algorithms as in Figure \ref{fig:SOFC_convergence}. The best solution represents a significant improvement of $31 \%$ increase in power and a $26 \%$ increase in efficiency compared to the original design reported in the literature \cite{radaideh2020efficient} using manual trial-and-error optimization based on expert knowledge. A comparison of our results to the literature is shown in Table \ref{tab:SOFC_power_eff}. These significant improvements in power output and efficiency show NEORL potential to optimize fuel cell designs for carbon-free energy production. 

\begin{table}[h]
\centering
\caption{\label{tab:SOFC_power_eff} A comparison of the optimal fuel cell power and efficiency calculated with different algorithms}
\begin{tabular}{cccccccc}
\toprule
Method & Power (W m$^{-2}$) & Efficiency\\
\hline
DE & $8236.1574$ & $46.64\%$\\ 
BAT & $8236.1574$ & $46.64\%$\\
GWO & $8232.3813$ & $46.62\%$\\
MFO & $8236.1574$ & $46.64\%$\\
PESA2 & $8236.1574$ & $46.64\%$\\
Original SOFC \cite{ni2007parametric,radaideh2020efficient} & $\approx 6300$ & $\approx 37.0 \%$\\
\bottomrule
\end{tabular}
\end{table}

Fuel cells have several benefits over conventional combustion-based technologies. Fuel cells as found in Table \ref{tab:SOFC_power_eff} can operate with efficiency close to 47\%, while most combustion engines provide a thermal efficiency of 20\%. The fuel cell efficiency can be even higher for other fuel cell designs, such as Alkaline fuel cells whose efficiency ranging from 60\%-70\%. In addition, fuel cells have zero carbon emissions compared to combustion engines. Hydrogen fuel cells like SOFC emit only water and feature no air pollutants that create smog and cause health problems.

\subsection{Optimizing Neural Networks for Nuclear Reactor Control}
\label{sec:mitr}

\begin{figure}
    \centering
    \includegraphics[width=\textwidth]{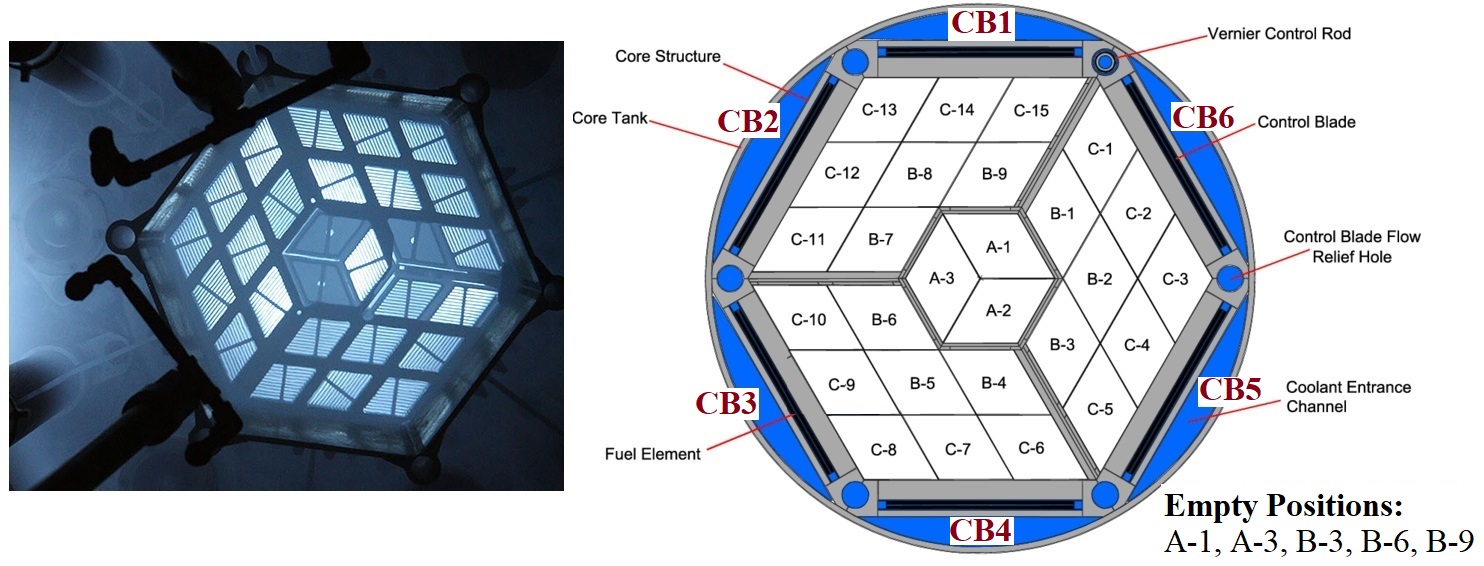}
    \caption{Top view of the MIT reactor core (Empty positions are used for experiments and do not have fuel)}
    \label{fig:mitr}
\end{figure}

The research on using machine and deep learning neural networks for nuclear reactor analysis is well-established. Examples of successful applications include using neural networks to predict nuclear power plant dynamic behaviour \cite{el2021artificial, bae2021real}, data-drive modeling of boiling heat transfer \cite{liu2018data}, nuclear multiphysics modeling \cite{radaideh2019combining}, uncertainty quantification of multiphase flow \cite{liu2021uncertainty}, and similar others. The Massachusetts Institute of Technology (MIT) research reactor was constructed in 1956 and upgraded in 1974. The reactor is a light-water cooled reactor with 6 MW thermal power. The primary applications of the reactor are research applications such as irradiation experiments to test nuclear materials for large nuclear power plants. Top view of the MIT reactor is shown in Figure \ref{fig:mitr}. The reactor core contains a total of 27 locations, where 5 locations are used for in-core experiments (A-1, A-3, B-3, B-6, B-9), and 22 locations are occupied by nuclear fuel elements. Each fuel element has a rhomboidal plate-type shape of highly enriched uranium oxide fuel in a dispersed aluminium matrix (see Figure \ref{fig:mitr}). The reactor fuel positions are surrounded by six control blades (CB), located around the outer periphery of the core, which are strong neutron absorbers used for fission reaction control. Controlling the insertion depth of each CB is very important for reactor control and stability. Therefore, in this problem, we use neural network to model the relationship between the CB height and the power produced by the 22 fuel elements. The work by Dave et al. \cite{dave2021empirical} explored the usage of machine learning methods such as gradient boosting and Gaussian process regression to generate empirical models for the MIT reactor. Therefore, we have used the proposed model of the MIT reactor developed by Dave et al. \cite{dave2021empirical} to generate 1000 samples of different CB depths and their corresponding power levels of the 22 fuel elements. These samples are used to train a neural network model to control the reactor. In a separate study \cite{dave2020thermal}, the authors reported about 1\%-2\% noise in power signals coming from the reactor sensors. Therefore, to account for label noise during training, we added Gaussian noise to 350 samples (1/3 of the dataset) with the prescribed noise range. The network is then trained based on both clean and noisy data, while it is tested to predict clean data. Given the small value of the signal noise, it is unlikely this will have large impact on the training process. 

The objective of this application is to \textit{use NEORL to build the neural network such that the mean absolute error (MAE) in the test set can be as low as possible}. In other words, the optimized neural network will predict the reactor power globally, given the positions of the six CBs. According to this paper \cite{song2020learning}, for labelled data with small noise, classical regularization methods such as dropout or weight decay (L1, L2) is sufficient to ensure model robustness. We have used 800 samples for training (contain clean and noisy data) and 200 samples for testing (with clean data only). We formulate the optimization problem as follows

\begin{equation}
\label{eq:mitr}
  \min_{\vec{x}} f(\vec{x}) = \min_{\vec{x}} \text{MAE}(\vec{x}) = \frac{1}{N_{test}}\sum_{i=1}^{N_{test}} [\hat{y}(\vec{x})-y]^2
\end{equation}
where $N_{test} = 200$ is the test set size, $\hat{y}$ is the neural network prediction, and $y$ is the target/real value from the test set. The input vector $\vec{x}$ represents the features of the feedforward neural network to be optimized by NEORL, which are:
\begin{enumerate}
    \item Number of hidden layers (Layers): A \textit{discrete} variable in the range [2,7]. 
    \item Learning rate (LR): A \textit{continuous} variable in the range [1e-4,1e-3]. 
    \item Batch size (Batch): A \textit{categorical} variable from the list \{8, 16, 32, 64, 128\}.
    \item Regularization for noise handling (Dropout): A \textit{continuous} variable in the range [0,0.5] to improve network robustness against noisy labels.  
    \item Number of nodes in layer 1 ($N_1$): A \textit{discrete} variable in the range [100,200].
    \item Number of nodes in layer 2 ($N_2$): A \textit{discrete} variable in the range [100,150].
    \item Number of nodes in layer 3 if used from item 1 ($N_3$): A \textit{discrete} variable in the range [100,150].
    \item Number of nodes in layer 4 if used from item 1 ($N_4$): A \textit{discrete} variable in the range [50,100].
    \item Number of nodes in layer 5 if used from item 1 ($N_5$): A \textit{discrete} variable in the range [50,100].
    \item Number of nodes in layer 6 if used from item 1 ($N_6$): A \textit{discrete} variable in the range [25,50].
    \item Number of nodes in layer 7 if used from item 1 ($N_7$): A \textit{discrete} variable in the range [25,50].
\end{enumerate}

Therefore, this application is a NEORL demonstration of mixed discrete/categorical/continuous optimization. We used five different NEORL algorithms to solve this problem: HHO, PESA, BAT, ES, and GWO, which all can solve mixed optimization problems. All algorithms have been executed for 10 generations, 10 individuals per generation, leading to a total of 100 neural network architectures evaluated per algorithm. Table \ref{tab:mitr_comp} lists the final results for all algorithms, which include the lowest MAE achieved and the optimized neural network variables corresponding to this MAE. We can notice very good agreement between the algorithms, especially between BAT and GWO, with four algorithms predicting with MAE $\leq 20$ Watt. All algorithms agree on the optimal network to be shallow with 2 hidden layers rather than being deep. In addition, all algorithms agree on batch size of 8, and on a minimal to zero dropout rate. This again confirms that the effect of the noise on the network performance is negligible such that dropout regularization is not needed. All tests have been executed using a workstation with 2 CPU nodes, 16 processors, 32 threads, 128 GB of memory, and GPU support.  

% Table generated by Excel2LaTeX from sheet 'control_tune'
\begin{table}[htbp]
  \centering
  \small
  \caption{Comparison of different neural network architectures for the MIT reactor as found by different NEORL optimizers}
    \begin{tabular}{lllllllllllll}
    \toprule
    Method & Layers & LR & Batch & Dropout & $N_1$ & $N_2$ & $N_3$ & $N_4$ & $N_5$ & $N_6$ & $N_7$ & MAE (Watt) \\
    \midrule
    BAT   & 2     & 5.64E-04 & 8     & 0.00  & 148   & 147   & 0     & 0     & 0     & 0     & 0     & 17.2 \\
    GWO   & 2     & 6.71E-04 & 8     & 0.00  & 102   & 137   & 0     & 0     & 0     & 0     & 0     & 17.8 \\
    ES    & 2     & 5.33E-04 & 8     & 0.00  & 184   & 126   & 0     & 0     & 0     & 0     & 0     & 18.3 \\
    PESA  & 2     & 8.63E-04 & 8     & 0.00  & 101   & 122   & 0     & 0     & 0     & 0     & 0     & 19.9 \\
    HHO   & 2     & 3.41E-04 & 8     & 0.06  & 184   & 123   & 0     & 0     & 0     & 0     & 0     & 26.4 \\
    \bottomrule
    \end{tabular}%
  \label{tab:mitr_comp}%
\end{table}%

In Table \ref{tab:mitr_others}, we benchmark BAT, GWO, and PESA as the best NEORL algorithms to other algorithms from nlopt \cite{johnson2014nlopt} and EvoloPy \cite{faris2016evolopy} in terms of lowest MAE and computing time. As indicated in Table \ref{tab:pycomp}, both nlopt and EvoloPy do not support parallel optimization, thus these algorithms have been executed on a single processor for same amount of function evaluations as NEORL in Table \ref{tab:mitr_comp}. We can notice how much time saving NEORL can achieve in Table \ref{tab:mitr_others}, as PESA (the fastest algorithm among all), can be 15 times faster compared to BOBYQA, and up to 18 times faster compared to sine cosine algorithm. Not only that, but also BAT and GWO are showing lower MAE results than other nlopt/EvoloPy algorithms.

% Table generated by Excel2LaTeX from sheet 'mitr_others'
\begin{table}[htbp]
  \centering
  \small
  \caption{Comparison of best NEORL algorithms against other frameworks for the MIT reactor neural network optimization}
  \begin{threeparttable}
    \begin{tabular}{lll}
    \toprule
    Algorithm (Framework) & MAE (Watt)   & \thead{Computational \\ Time (s)} \\
    \midrule
    Controlled Random Search (nlopt) \cite{johnson2014nlopt} & 19.0 & 912 \\
    Nelder Mead Simplex (nlopt) \cite{johnson2014nlopt} & 19.2 & 871 \\
    BOBYQA (nlopt) \cite{johnson2014nlopt} & 17.9 & 833 \\
    Firefly Algorithm (EvoloPy) \cite{faris2016evolopy} & 42.7 & 950 \\
    Multi-verse Optimizer  (EvoloPy) \cite{faris2016evolopy} & 33.5 & 927 \\
    Since Cosine Algorithm (EvoloPy) \cite{faris2016evolopy} & 18.9 & 981 \\
    PESA (NEORL) & 19.9 & 55* \\
    GWO (NEORL) & 17.8 & 95* \\
    BAT (NEORL) & 17.2 & 226* \\
    \bottomrule
    \end{tabular}%
    \begin{tablenotes}
      \small
      \item $*$ PESA, BAT, and GWO are running in parallel with 10 processors per algorithm.   
    \end{tablenotes}
    \end{threeparttable}
  \label{tab:mitr_others}%
\end{table}%

\subsection{Additional Applications}

For brevity, we refer the reader to more NEORL applications available on the webpage\footnote{\url{https://neorl.readthedocs.io/en/latest/index.html}}.
We illustrate how RL algorithms can solve classical combinatorial problems such as travel salesman problem\footnote{\url{https://neorl.readthedocs.io/en/latest/examples/ex1.html}} and knapsack problem\footnote{\url{https://neorl.readthedocs.io/en/latest/examples/ex10.html}}, while NEORL hyperparameter tuning tools help RL algorithms to achieve an optimal performance. In addition, we have demonstrated how variety of classical and modern evolutionary algorithms can be used to design a three-bar truss\footnote{\url{https://neorl.readthedocs.io/en/latest/examples/ex6.html}}, welded beam\footnote{\url{https://neorl.readthedocs.io/en/latest/examples/ex3.html}}, pressure vessel\footnote{\url{https://neorl.readthedocs.io/en/latest/examples/ex8.html}}, and others, which involve solving different problem types with different constraints. Our results in this work, which come from a single framework, show very competitive performance with the literature results that include a very diverse range of algorithms and implementations.

\section{Conclusions}
\label{sec:conc}

In this work, we have presented NEORL as an advanced optimization framework that features different computational capabilities to solve large-scale optimization problems along with a friendly interface. However, we should emphasize that the novelty of this work is not only framework development with demonstration on engineering applications, but more importantly on the algorithm methodology being developed. The ability to bring the state-of-the-art algorithms from neural networks, evolutionary computation, and neuroevolution in a cohesive environment is the core contribution of this work. Furthermore, some of these algorithms are actually proposed and developed by the authors.

Featuring +25 algorithms from different categories in NEORL is vital to accommodate the ``No Free Lunch Theorems for Optimization'' \cite{wolpert1997no}, which state that ``for any optimization algorithm, any elevated performance over one class of problems is offset by performance over another class'', which would imply that any optimization algorithm performance averaged over all possible problems will be close to random search. Therefore, providing the ability to deploy variety of algorithms with different natures from neural, evolutionary, and neuroevolution categories on the same problem, can provide the analyst of more comprehensive results than relying on a single algorithm. 

%We basically have demonstrated that in \textbf{Supplementary Materials} section 1, when we have used 10 NEORL algorithms to solve more than 20 benchmark problems. The results in \textbf{Supplementary Materials} Table 1 are literally showing that there is no clear winner among the 10 algorithms, but the overall NEORL ensemble was able to achieve very satisfactory results over all benchmark problems. 

In conclusion, solving optimization problems is at the heart of many disciplines to find the most optimal solutions to their problems, which require advanced algorithms and friendly framework implementation. In this work, we proposed and benchmarked NEORL, which is a framework for NeuroEvolution Optimization with Reinforcement Learning. NEORL was developed with a main purpose to resolve the issues facing other Python optimization frameworks by providing an interface of state-of-the-art algorithms in the field of evolutionary computation, neural networks, reinforcement learning, and hybrid neuroevolution algorithms. The novelty of NEORL originates from the diversity of the implemented algorithms, advanced implementation, user-friendly interface, parallel computing support, hyperparameter tuning, detailed documentation, and demonstration of applications in mathematical and real-world engineering optimization. NEORL was benchmarked against many algorithms from the literature and against other popular optimization frameworks (nlopt, EvoloPy, DEAP, PyOpt); showing an excellent and competitive performance. 

Future steps of NEORL will focus on two major paths. First, we will add special algorithms and capabilities for direct multi-objective optimization if the user prefers to avoid using priori conversion methods. Second, more advanced applications will be pursued by the authors, in which NEORL is utilized directly to solve large-scale design problems with more problem-dependent physics guidance, which can be obtained in form of sensitivity coefficients \cite{borowiec2020validation, price2021methodology} or data-driven physical closures \cite{borowiec2021comprehensive}. Given its open-source nature, the authors also welcome contributions from researchers around the world who are willing to enrich NEORL family with new algorithms and/or applications.

\section{Data Availability}
\label{sec:avail}

NEORL framework is open-source, and it can be accessed from: \href{https://github.com/mradaideh/neorl}{https://github.com/mradaideh/neorl}. The documentation and tutorials are available here: \href{https://neorl.readthedocs.io/en/latest/index.html}{https://neorl.readthedocs.io/en/latest/index.html}.

All scripts used in this paper can be found under the \code{examples} directory within NEORL: \href{https://github.com/mradaideh/neorl/tree/master/examples/journal\_tests}{https://github.com/mradaideh/neorl/tree/master/examples/journal\_tests}.

\section*{Acknowledgment}

This work is sponsored by Exelon Corporation, a nuclear electric power generation company, under the award (40008739). The authors would like to thank Benoit Forget of Massachusetts Institute of Technology (MIT), Isaac Wolverton and Joshua Joseph of MIT Quest for Intelligence, and James J. Tusar and Ugi Otgonbaatar of Exelon corporation, for their feedback and discussions in the early phase of the project. 

\section*{Credit Author Statement}

\noindent \textbf{Majdi I. Radaideh}:  Conceptualization, Methodology, Software, Validation, Investigation, Data curation, Visualisation, Formal analysis, Writing - Original Draft. \\
\textbf{Katelin Du}: Methodology, Software, Validation, Formal analysis, Writing - Original Draft. \\
\textbf{Paul Seurin}: Methodology, Software, Validation, Formal analysis, Writing - Original Draft. \\
\textbf{Devin Seyler}: Methodology, Software, Validation, Formal analysis, Writing - Original Draft. \\
\textbf{Xubo Gu}: Methodology, Software, Writing – Review and Edit. \\
\textbf{Haijia Wang}: Methodology, Software, Writing – Review and Edit. \\
\textbf{Koroush Shirvan}: Conceptualization, Methodology, Funding acquisition, Resources, Writing – Review and Edit.

%\section*{References}
\bibliographystyle{elsarticle-num}
%\bibliographystyle{apa}
%\biboptions{authoryear}
{
\footnotesize \bibliography{references}}

\end{document}